\documentclass{article}

\usepackage[numbers,sort&compress]{natbib}
\usepackage[main, final]{neurips_2026}

\usepackage[utf8]{inputenc} % allow utf-8 input
\usepackage[T1]{fontenc}    % use 8-bit T1 fonts
\usepackage{hyperref}       % hyperlinks
\usepackage{url}            % simple URL typesetting
\usepackage{booktabs}       % professional-quality tables
\usepackage{amsfonts}       % blackboard math symbols
\usepackage{nicefrac}       % compact symbols for 1/2, etc.
\usepackage{microtype}      % microtypography
\usepackage{xcolor}         % colors
\usepackage{graphicx}
\usepackage{wrapfig}
\usepackage{multirow}
\usepackage[most]{tcolorbox}
\usepackage{xcolor}
\usepackage{tikz}
\usepackage{enumitem}

\definecolor{stepblue}{HTML}{3B82F6}
\definecolor{stepbluefill}{HTML}{EFF6FF}

\newcommand{\stepcircle}[1]{%
  \tikz[baseline=(char.base)]{
    \node[
      circle,
      draw=stepblue,
      fill=stepbluefill,
      text=black,
      line width=1.1pt,
      minimum size=1.65em,
      inner sep=0pt,
      font=\bfseries\small
    ] (char) {#1};
  }%
}

\newtcolorbox{promptbox}[2][]{
  enhanced,
  breakable,
  colback=gray!3,
  colframe=gray!45,
  boxrule=0.4pt,
  arc=1mm,
  left=1.2mm,
  right=1.2mm,
  top=1mm,
  bottom=1mm,
  fonttitle=\bfseries,
  coltitle=black,
  colbacktitle=gray!12,
  title=#2,
  #1
}
\newcounter{prompt}

\newcommand{\mingyue}[1]{\textcolor{black}{#1}} %purple
\title{Intent2CAD: How Semantic--Parametric Supervision Shapes Text-to-CAD Generation}

\author{%
  Jinbo Yang\\
  Beijing Institute of Technology\\
  Beijing, China \\
  \texttt{jinboy@bit.edu.cn} \\
  \And
  Mingyue Yuan\thanks{Corresponding author (mingyue.yuan@unsw.edu.au)}  \\
  University of New South Wales \\
  Sydney NSW, Australia  \\
  \texttt{mingyue.yuan@unsw.edu.au} \\
  \And
  Shikai Jing\thanks{Corresponding author (jingshikai@bit.edu.cn)} \\
  Beijing Institute of Technology \\
  Beijing, China  \\
  \texttt{jingshikai@bit.edu.cn} \\
}

\begin{document}

\maketitle

\begin{abstract}
Text-to-CAD generation aims to make CAD modeling more accessible by generating executable CAD programs from natural-language descriptions. However, practical CAD workflows require more than executable geometry: generated programs should also expose reusable structure that supports later modification. We introduce \emph{intent-oriented text-to-CAD generation}, where design intent is expressed as recoverable code-level signals: semantic features, reusable parameters, and lightweight parametric relations. Rather than changing the model architecture or decoding procedure, we investigate how supervision representation granularity affects both CAD generation and zero-shot intent-preserving editing. To enable this study, we construct \emph{IntentCAD-100K}, a 100K paired text--code dataset built by converting low-level CAD construction traces into executable CadQuery programs. We further construct \emph{IntentCAD-Edit-1K} to evaluate zero-shot intent-preserving editing from the same representation perspective. Intent2CAD uses a two-stage conservative lifting pipeline: semantic lifting rewrites reliable primitive groups into feature-level operations, while parametric lifting exposes lightweight parametric relations. Under controlled fine-tuning, semantic lifting mainly improves generation reliability and feature recovery, reducing IR from 2.39\% to 0.22\% and achieving 99.96\% API Recovery, while parametric lifting makes reusable relation structure explicit and further improves relation-aware zero-shot editing. 
Datasets and code are available at \url{https://github.com/JinboYang94/Intent2CAD}.

\end{abstract}

\section{Introduction}
\label{sec:intro}

Computer-Aided Design (CAD) is a fundamental tool in modern engineering design and manufacturing, where designers create, refine, and iterate product designs throughout the development process~\cite{zhang2026large,heidari2024geometric,willis2021fusion}. Because traditional CAD workflows remain time-consuming and expertise-intensive~\cite{robertson1993cad}, recent text-to-CAD research has explored how Large Language Models (LLMs) can generate CAD scripts from natural language descriptions~\cite{khan2024text2cad,govindarajan2025cadmium,wang2025cadfusion,guan2025cadcoder,xie2025text,li2025cadllama}. This direction is promising, but practical CAD modeling is not only about producing a final shape. In parametric CAD modeling, a generated CAD program should support later reuse and modification, not just match the final shape.

Most existing text-to-CAD studies therefore leave an important gap. They primarily focus on geometric fidelity and program executability~\cite{khan2024text2cad,govindarajan2025cadmium,wang2025cadfusion,guan2025cadcoder,xie2025text}. These criteria are necessary, but they do not fully capture the needs of parametric CAD workflows~\cite{otey2018revisiting,kyratzi2022integrated,casey2025aligning}. 
A program that approximates the target geometry with many low-level primitives may pass a shape metric, while still making a simple later edit, such as changing a shared radius or preserving symmetry, difficult to perform consistently. 
In this paper, we call this broader requirement \emph{intent-oriented text-to-CAD generation}: the generated program should not only reproduce the requested shape, but also expose recoverable code-level intent signals that support reuse and local refinement.

We use \emph{intent} in a restricted sense, referring only to recoverable code-level signals such as supported semantic features, reusable parameters, and lightweight parametric relations, rather than full design history or constraint-solving processes~\cite{bouma1995geometric}.
The \textbf{supervision representation} is important to this problem. Most current benchmarks are built from low-level construction traces composed of primitives such as lines, arcs, and circles~\cite{wu2021deepcad,khan2024text2cad,govindarajan2025cadmium,wang2025cadgpt}. While expressive, these traces can hide the semantic and parametric structure that makes CAD programs reusable and editable.

\begin{wrapfigure}{r}{0.35\textwidth}
    \centering
    \includegraphics[width=\linewidth]{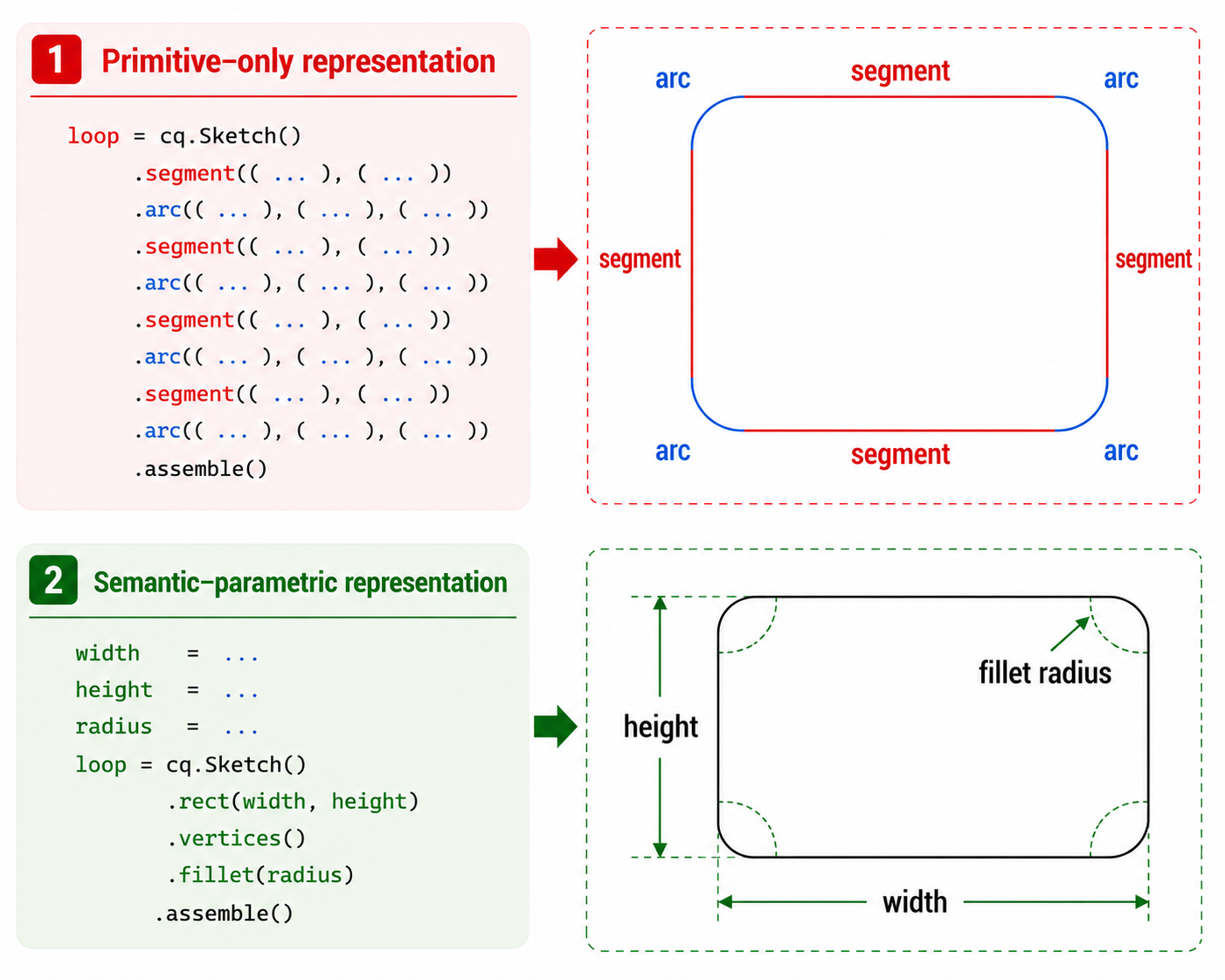}
    \caption{Illustration of the representation gap between primitive-only and semantic--parametric CAD supervision.}
    \label{fig:code_compare}
\end{wrapfigure}

Figure~\ref{fig:code_compare} illustrates this representation gap. Both forms can describe same geometry, but they expose different supervision signals: the primitive-only form emphasizes low-level primitives, whereas the semantic--parametric form makes high-level feature and reusable parameters explicit in code. This motivates our main question: when the model, training objective, text descriptions, and inference setup are fixed, how much does the target representation itself affect text-to-CAD generation and intent-preserving editing?

To answer this question, we introduce \emph{Intent2CAD}, a semantic--parametric supervision representation implemented as executable CadQuery code. 
Following a conservative \textbf{\textit{``lift when reliable, preserve when uncertain''}} rule, Intent2CAD rewrites supported primitive groups into high-level CAD features when geometric and constraint evidence agree, while retaining low-level fallbacks for unsupported or ambiguous geometry. 
It exposes four code-level representation elements: semantic features, explicit parameters, parametric relations, and fallbacks.

We construct this representation through a two-stage lift-and-validate pipeline. 
Stage I performs tag-preserving reconstruction and constraint-grounded semantic lifting, rewriting reliable primitive groups into feature-level CadQuery operations with explicit parameters. 
Stage II performs restricted relation induction by local parameter rewriting, exposing shared variables and simple derived expressions when they pass execution, geometric consistency, and locality checks. 
The pipeline produces three supervision granularities for study: primitive-only, semantic, and semantic--parametric supervision.

Using this pipeline, we build \emph{IntentCAD-100K}, a 100K paired text--code dataset derived from raw DeepCAD~\cite{wu2021deepcad} construction sequences. 
Beyond standard Invalid Ratio (IR) and Chamfer Distance (CD), we evaluate whether generated code recovers semantic features, preserves shared/equal parameters, and preserves derived parametric relations.
We further construct \emph{IntentCAD-Edit-1K}, a zero-shot intent-preserving editing benchmark with 1K aligned Feature Edit (FE), Consistency-preserving Edit (CE), and Derived-Relation Edit (DE) samples. 
Unlike prior text-based CAD editing work that trains specialized editors~\cite{yuan2025cad,zhang2025geocad,zhang2024flexcad}, we use the same text-to-CAD generation checkpoints to evaluate local refinement while preserving representation-specific intent signals. We also evaluate visually grounded editing to examine target localization capacity with spatial descriptions.

Experiments show that representation granularity has measurable effects.  
Under the same supervised fine-tuning setup, semantic lifting reduces IR from 2.39\% to 0.22\%, lowers mean CD from 1.44 to 0.46 ($\times 10^3$), and achieves 99.96\% API Recovery. 
Parametric lifting keeps reliability and typical geometry comparable to semantic supervision, while recovering relation structure with 93.35\% ERS and 95.25\% PRS. 
In zero-shot editing, semantic--parametric supervision further improves relation-aware edits, raising CE@ES from 91.00\% to 95.82\% and DE@ES from 77.09\% to 88.73\%. Our contributions can be summarized as follows:

\begin{itemize}[leftmargin=*]
     \item We introduce \emph{intent-oriented text-to-CAD generation} as a representation-centered formulation that goes beyond executable geometry by focusing on recoverable semantic features, reusable parameters, and lightweight parametric relations.
    \item We construct \emph{IntentCAD-100K}, a large-scale aligned text--code dataset that converts low-level CAD construction traces into primitive-only, semantic, and semantic--parametric executable CadQuery targets through our automated conservative lifting-and-validation pipeline.
    \item We construct \emph{IntentCAD-Edit-1K} as a validation benchmark for studying how representation granularity affects zero-shot intent-preserving editing, covering feature edits, consistency-preserving edits, and derived-relation edits without edit-specific training.
    \item Experiments show that semantic lifting improves generation reliability and feature-level editability, while parametric lifting improves relation preservation and relation-aware editing.
\end{itemize}

\vspace{-1em}
\section{Related Work}
\vspace{-0.5em}
\subsection{Symbolic CAD Representations for Text-to-CAD}

Text-to-CAD differs from general shape generation~\cite{wu2016learning,park2019deepsdf,nash2020polygen,yang2019pointflow} because the symbolic output should be an editable construction program, not only a final geometry. 
Prior work has therefore adopted symbolic CAD targets, including sketch-and-extrude sequences, primitive and constraint graphs, domain-specific CAD formats, and executable CAD code~\cite{wu2021deepcad,seff2020sketchgraphs,khan2024text2cad,govindarajan2025cadmium,guan2025cadcoder,xie2025text,willis2021engineering}. 
Early approaches represent CAD models as low-level construction traces over primitives such as lines, arcs, circles, and extrusion commands~\cite{wu2021deepcad,seff2020sketchgraphs}, while recent text-conditioned systems extend these targets to natural-language generation and executable CadQuery code~\cite{khan2024text2cad,govindarajan2025cadmium,guan2025cadcoder,xie2025text}. 
However, a code target alone does not determine the semantic quality of the representation. 
Existing supervision is still largely derived from primitive-level traces, which scale well but leave feature grouping, editable variables, and geometric or parametric relations implicit. 
This limits their ability to expose the modeling structure needed for refinement and reuse. 
Our work addresses this representation gap by constructing aligned primitive-only, semantic, and semantic--parametric targets.

\vspace{-0.5em}
\subsection{Design Intent in Parametric CAD Modeling}
\vspace{-0.5em}
Design intent in parametric CAD modeling describes how a model is expected to behave under modification~\cite{otey2018revisiting,kyratzi2022integrated}. 
A geometrically valid model may still be poorly editable if feature structure, dimensions, or relations are lost~\cite{casey2025aligning}. 
Prior work has studied such relational structure in constraint-based sketches~\cite{hoffmann2005brief,yang2022discovering}, including primitive--constraint graph representations~\cite{seff2020sketchgraphs} and constrained sketch generation with automatic constraint completion~\cite{para2021sketchgen}.
Our work extends this editability perspective to code-based parametric 3D CAD. 
Rather than reconstructing a full constraint system or using solver feedback, we expose recoverable code-level signals, such as semantic features, reusable parameters, and lightweight parametric relations, directly in executable CAD programs.

\vspace{-0.5em}
\subsection{LLM-based Text-to-CAD Generation and Editing}
\vspace{-0.5em}

LLMs have recently been applied to text-to-CAD through sequential command prediction~\cite{khan2024text2cad,govindarajan2025cadmium}, B-rep-grounded command generation with pointer-based entity selection~\cite{qi2026pointer}, multimodal generation with visual feedback~\cite{wang2025cadfusion}, and direct generation of executable CAD code~\cite{xie2025text,guan2025cadcoder,li2025cadllama}. 
Parallel efforts study controllable generation and text-guided editing, including hierarchy-aware control~\cite{zhang2024flexcad}, local geometry control~\cite{zhang2025geocad}, locate-then-infill editing~\cite{yuan2025cad}, and unified CAD prototyping with dedicated tokenization~\cite{wang2025cad}.
CAD-Tokenizer~\cite{wang2025cad} learns compact, primitive-aware tokens for generation and editing, while cadrille~\cite{kolodiazhnyi2025cadrille} focuses on multimodal CAD reconstruction from text, images, and point clouds.
However, existing work rarely studies the semantic and parametric granularity of the supervision target itself. 
Execution and geometric checks verify whether a CAD program runs and matches the target shape, but do not ensure reusable feature semantics or meaningful parametric relations. 
Rather than proposing another editor or decoding pipeline, our work focuses on the upstream target representation and studies whether explicit semantic--parametric supervision improves learning and editability under standard supervised fine-tuning.

\vspace{-0.5em}
\section{Method: Intent-Explicit Supervision for Text-to-CAD}
\label{methods}

\vspace{-0.5em}
\subsection{Intent2CAD: Intent-Explicit CadQuery Representation}
\label{sec:hybrid_representation}
\vspace{-0.5em}

Intent2CAD provides semantic–-parametric representation supervision for intent-oriented text-to-CAD generation. Rather than changing the model architecture, training objective, or decoding procedure, it changes the \textbf{representation} of the target code that the model is trained to produce. 
In this way, each CAD program is represented as executable CadQuery code in which recoverable design intent is made \textbf{explicit}: semantic features expose \textbf{reusable CAD modeling operations}, explicit parameters name \textbf{editable design variables}, and lightweight parametric relations capture simple \textbf{geometric and parameter relations}, while low-level fallbacks retain coverage for complex \textbf{geometry}. This allows standard supervised fine-tuning to learn not only the resulting executable geometry, but also how that geometry is constructed and parameterized.

\begin{figure}[h]
    \centering
    \includegraphics[width=1\textwidth]{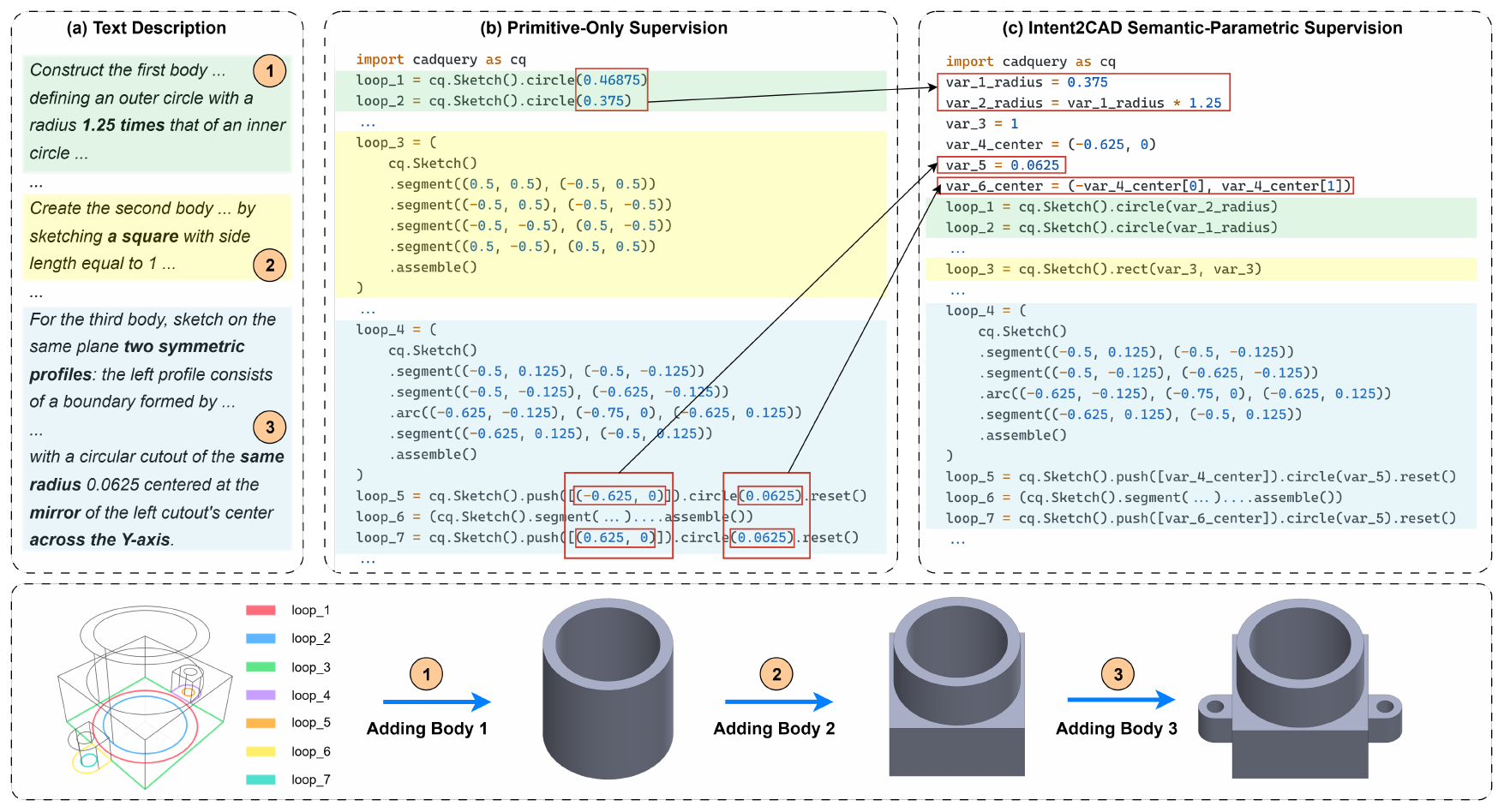}
    \caption{
    Illustration of the Intent2CAD semantic--parametric representation.
    (a) Intent-oriented text description of a CAD model.
    (b) Corresponding program using primitive-only supervision.
    (c) Corresponding program using semantic-parametric supervision.
    The bottom block shows the modeling process for each step.
    Red boxes and arrows indicate parameters and relations exposed by Intent2CAD.
    }
    \label{fig:intent2cad_rep}
\end{figure}

By contrasting the two supervision representations of the same CAD instance, Figure~\ref{fig:intent2cad_rep} illustrates how Intent2CAD changes the supervision target. 
The primitive-only target keeps the recovered construction as low-level primitives, where semantic structures appear as segments, arcs, and repeated numeric constants. 
In the semantic--parametric target, recoverable primitive groups are rewritten into semantic CAD operations such as \texttt{rect}, editable attributes are named as explicit parameters such as \texttt{radii}, \texttt{widths}, \texttt{heights}, and \texttt{centers}, and related values are represented through shared variables or simple derived expressions. 
For example, proportional radii and symmetric centers are exposed as code-level relations, as indicated by the red boxes and arrows. 
Unsupported primitive groups remain as low-level fallback loops to preserve executable coverage for complex geometry. 
The intermediate semantic target used in the controlled study stops before relation lifting: it keeps the semantic feature rewrites, but lacks explicit parameters and relations.

We define the supported operations based on common CadQuery modeling patterns that can be reliably recovered from low-level sketch--extrude traces using geometric checks and source constraint evidence. 
This conservative set includes semantic CAD construction commands, such as \texttt{rect}, \texttt{regular polygon}, \texttt{irregular polygon}, \texttt{slot}, \texttt{fillet}, and \texttt{array}, as well as lightweight sketch relation cues, including \texttt{PARALLEL}, \texttt{HORIZONTAL}, \texttt{VERTICAL}, \texttt{PERPENDICULAR}, and \texttt{TANGENT}. 
To make repeated or related values more editable than repeated numeric constants, Intent2CAD further introduces a restricted set of recoverable code-level relation signals, including \texttt{EQUAL}, \texttt{CONCENTRIC}, \texttt{MIRROR}, \texttt{DISTANCE}, and common \texttt{RATIO} relations between dimensional parameters. 
Equal lengths, radii, or centers are represented by shared variables, while mirrored or distance-based structures are represented through symmetric or offset parameter definitions. 
These relations are not a full constraint system; they expose recurring geometric and parametric regularities in code.

\vspace{-0.5em}
\subsection{Lift-and-Validate Construction Pipeline}
\label{sec:conversion_pipeline}
\vspace{-0.5em}

To construct Intent2CAD supervision at scale, we introduce an automatic \textbf{lift-and-validate construction pipeline} that converts each \textbf{raw CAD sample} into a paired \emph{text--code} instance. The pipeline combines deterministic geometric checks, recovered constraint evidence, constrained LLM-based local rewriting, and executable validation, without requiring manual annotation of semantic features or parameter relations. Candidate lifts or rewrites are accepted only when their reliability can be established; otherwise, the original low-level implementation is preserved. This makes the construction process scalable while keeping the generated target code executable and geometrically faithful.

\begin{figure}[htbp]
    \centering
    \includegraphics[width=1\textwidth]{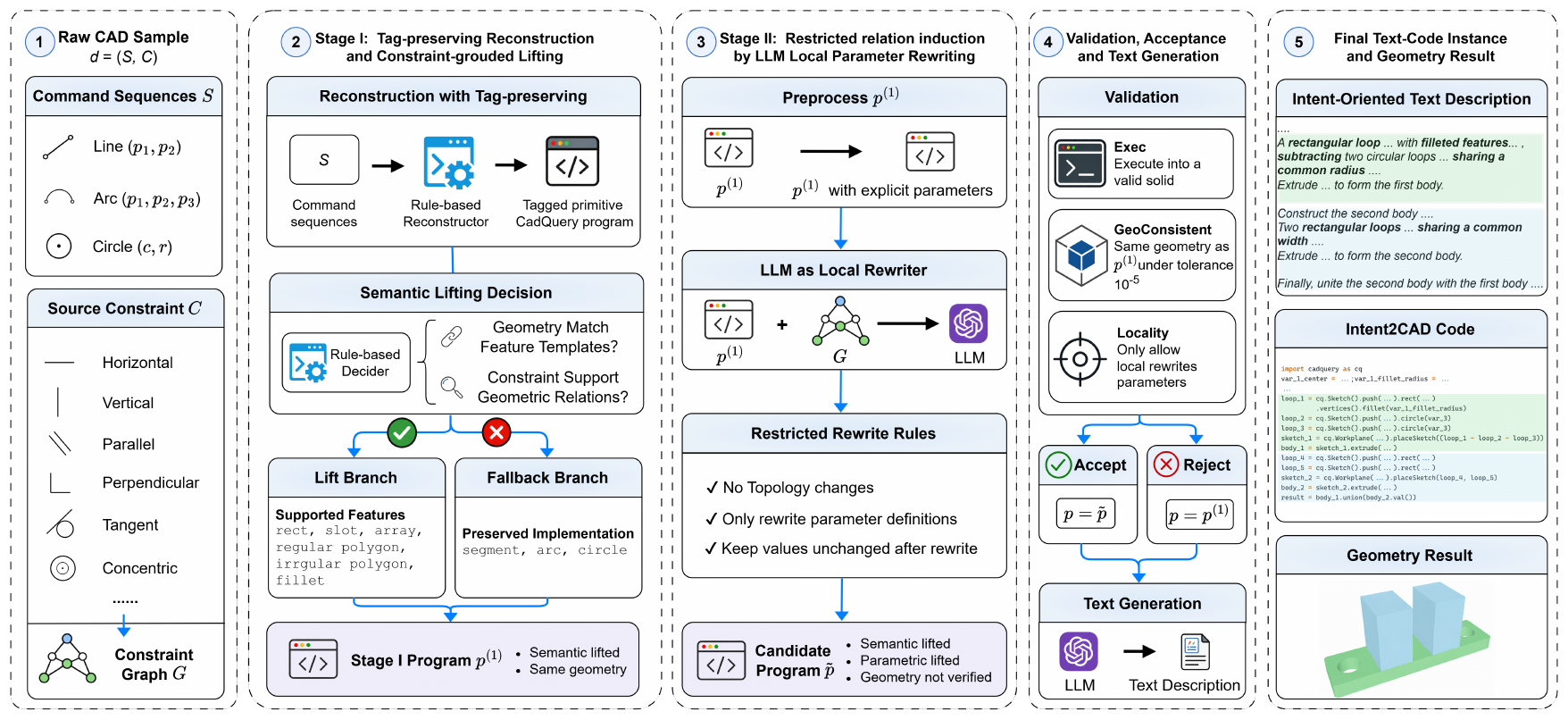}
    \caption{Intent2CAD lift-and-validate construction pipeline. 
    In \protect\stepcircle{5}, green indicates the rounded-rectangle base, and blue indicates two same-width rectangular extrusions.
    }
    \label{fig:pipeline}
\end{figure}

\vspace{-0.5em}
As shown in Figure~\ref{fig:pipeline} \stepcircle{1}, each \textbf{raw CAD sample} is represented as $d=(S,C)$, where \textbf{command sequences} $S$ contain the original low-level sequential CAD commands and \textbf{source constraint} $C$ contains recovered geometric and dimensional relations. 
We extract a \textbf{constraint graph} $G$ from $C$ and use it as structural evidence for reliable semantic lifting and safe exposure of parameter relations, rather than to reconstruct the full design history.
The intermediate programs are retained, yielding \mingyue{three aligned supervision granularities}: the primitive reconstruction before lifting, the Stage-I semantic program, and the Stage-II semantic--parametric program.

\vspace{-1em}
\paragraph{Stage I: Tag-preserving reconstruction and constraint-grounded lifting.}
As shown in Figure~\ref{fig:pipeline} \stepcircle{2}, \textbf{Stage I} first performs \textbf{tag-preserving reconstruction}: we reconstruct an executable low-level CadQuery program from Command Sequences $S$ using a rule-based reconstructor, while assigning each low-level primitive a unique tag identifier. 
The recovered constraints $C$ are represented as a typed relation graph $G$, whose nodes are tagged geometric entities and whose edges encode available local geometric or dimensional relations. 
This yields a \textbf{tagged primitive CadQuery program} for checking candidate primitive groups against geometry and recovered relations.

Semantic lifting is then applied through a conservative \textbf{semantic lifting decision}. 
For each candidate primitive group $g$ and supported feature template $\tau$, lifting is accepted only if
\begin{equation}
\mathrm{Liftable}(g,\tau)=
\mathbf{1}\!\left[
\mathrm{GeomMatch}(g,\tau)\land
\mathrm{ConstraintSupport}(G[g],\tau)
\right].
\end{equation}
where $\mathrm{GeomMatch}$ checks whether the local primitive arrangement matches the feature template, while $\mathrm{ConstraintSupport}$ checks whether the available relations in $G[g]$ support the corresponding geometric relations. 
Accepted groups enter the \textbf{lift branch} and are rewritten into a feature-level expression with explicit parameters; rejected or ambiguous groups enter the \textbf{fallback branch}, where the original implementation is preserved.

\paragraph{Stage II: Restricted relation induction by local parameter rewriting.}
Stage I makes feature structure explicit, but reusable relations may still remain implicit in parameter expressions, such as repeated dimensions, equality-style relations, shared centers, and simple derived forms. 
As shown in Figure~\ref{fig:pipeline} \stepcircle{3}, Stage II begins by \textbf{preprocessing} the Stage-I program $p^{(1)}$ into $p^{(1)}$ with explicit parameters, so that candidate parameter definitions and use sites can be rewritten locally. 
The LLM is then used only as a constrained \textbf{local rewriter}: it receives $p^{(1)}$ and the relation graph $G$, and predicts a structured rewrite plan $\phi$ over parameter definitions and parameter-use sites. 
The plan is applied deterministically to obtain a candidate relation-enhanced program $\tilde{p}$.
In our \textbf{implementation}, we use GPT-5.4 with a structured JSON output format; the full prompt is provided in Appendix~A. 
The LLM output is not directly used as supervision, but is treated as a rewrite proposal that must follow \textbf{restricted rewrite rules} and pass deterministic validation before acceptance. 
Specifically, the rewrite interface is limited to a whitelist of parameter definitions and parameter-use sites: it cannot change topology, cannot introduce new geometric operations or alter their order, and can only rewrite parameter expressions into shared variables or simple derived forms while keeping evaluated values unchanged. 
Applying this proposal produces a candidate program $\tilde{p}$ that is semantically and parametrically lifted, but whose geometry has not yet been verified.

\paragraph{Validation, acceptance and text generation.}
As shown in Figure~\ref{fig:pipeline} \stepcircle{4},
LLM-proposed rewrites are treated as candidates rather than directly accepted, and we apply deterministic \textbf{validation} to every candidate program. 
A candidate rewrite is accepted only if
\begin{equation}
\mathrm{Valid}(\tilde{p}, p^{(1)}) =
\mathbf{1}\left[
\mathrm{Exec}(\tilde{p}) \wedge
\mathrm{GeoConsistent}(\tilde{p}, p^{(1)}) \wedge
\mathrm{Locality}(\phi)
\right],
\end{equation}
where $\mathrm{Exec}$ checks whether the rewritten CadQuery program executes into a valid solid; $\mathrm{GeoConsistent}$ checks whether the rewritten program remains geometrically consistent with the Stage-I program under a fixed geometric tolerance; and $\mathrm{Locality}$ verifies that the rewrite only changes allowed parameter definitions or parameter uses, without changing the program skeleton. The final accepted program is $\tilde{p}$ if validation succeeds, and $p^{(1)}$ otherwise. This fallback mechanism allows the pipeline to process large collections automatically, since uncertain relation rewrites are rejected rather than manually repaired.

After the final program $p$ is determined, we use GPT-5.4 to generate its paired text description $x$ from the \textbf{acceptance} procedure. The \textbf{text descriptions} are written at the design-intent level, including supported feature semantics and, when applicable, simple geometry or parameter relations. To avoid representation-specific leakage, the descriptions do not expose implementation-specific function names, variable names, or the chosen target representation.

Therefore, the pipeline yields three granularities for the controlled study: primitive-only supervision, semantic supervision after Stage I, and semantic--parametric supervision after Stage II. These forms share the same underlying geometry and text description, but differ in the symbolic structure exposed in the target code. Thus, the controlled study changes the target supervision representation while keeping the input description and underlying geometry fixed. Overall, this automatic supervision-construction process produces validated target representations at scale.

\vspace{-1em}
\section{Experiments}
\label{experiments}
\vspace{-1em}

\subsection{Experimental Setup}
\label{sec:exp_setup}

\paragraph{IntentCAD-100K Construction and Splits.}
We evaluate Intent2CAD on IntentCAD-100K, which is constructed by applying the conversion pipeline in Section~\ref{sec:conversion_pipeline} to 178,238 raw construction sequences from DeepCAD~\cite{wu2021deepcad}. 
We retain samples that can be converted into executable CadQuery programs, normalized to a canonical coordinate scale, and passed through geometric validity checks; samples with unsupported operations or unrecoverable sketch/extrusion structure are discarded. 
Before splitting, we also remove duplicate primitive reconstructions using a quantized construction signature computed before semantic and parametric lifting, ensuring that the three supervision representations remain aligned after deduplication and reducing leakage from duplicate construction records. 
The final benchmark contains 100K paired text--code samples, randomly split into training, validation, and test sets with a ratio of 80/10/10. Appendix~A reports the data distribution, conversion pipeline, and duplicate-removal details.

\vspace{-0.5em}
\paragraph{Statistical Analysis.}
We apply bootstrap analysis to the controlled representation comparisons in Section~\ref{sec:representation_ablation} and Section~\ref{sec:editability}. The preliminary baseline comparison is reported as a diagnostic reference. For these controlled comparisons, we report 95\% percentile bootstrap confidence intervals using 10,000 resamples over test instances. For representation comparisons, we additionally report paired bootstrap confidence intervals in Appendix D for metric differences by resampling the same instance indices across representations. Positive differences indicate that the main representation performs better: for lower-is-better metrics such as IR and CD, we compute baseline minus main; for higher-is-better metrics, we compute main minus baseline.

\vspace{-0.5em}
\paragraph{Representation setup.}
We evaluate three aligned supervision representations: primitive-only supervision ($-$), semantic supervision ($S.$), and semantic--parametric supervision ($S.{+}P.$). 
The controlled comparisons use the same text instructions, data split, model backbone, training strategy, and inference setting, making representation granularity the primary varying factor. 

\vspace{-0.5em}
\paragraph{IntentCAD-Edit-1K Construction.}
We synthesize editing triples $(c_{\mathrm{src}}, e, c_{\mathrm{tgt}})$ from the $S.+P.$ part of IntentCAD-100K test set, where $c_{\mathrm{src}}$ is the source program, $e$ is an intent-oriented edit instruction, and $c_{\mathrm{tgt}}$ is the target edited program. 
Each sample contains a single edit intent, but the corresponding code change may involve multiple locations when required by a design relation. 
For each edit, the same instruction is paired with aligned primitive-only, semantic-only, and semantic--parametric source/target programs, so that the comparison isolates the effect of representation.

We construct three edit types that directly correspond to the intent-bearing structures exposed by our representation.
\textbf{(1) Feature Edit (FE):} replace a semantic feature or modify its attributes while preserving the remaining relations.
\textbf{(2) Consistency Edit (CE):} change one editable value while preserving equality-like relations encoded by shared variables or shared geometric references, such as \texttt{EQUAL} dimensions or \texttt{CONCENTRIC} centers.
\textbf{(3) Derived-Relation Edit (DE):} change one editable value while preserving lightweight derived relations encoded in the target program, such as \texttt{MIRROR} coordinates, \texttt{RATIO} dimensions, or \texttt{DISTANCE} offsets.
The final benchmark contains 1K samples, including 414 FE, 311 CE, and 275 DE samples.
Additional construction details and edit-type statistics are provided in Appendix C.

\vspace{-0.5em}
\paragraph{Evaluation Metrics.}

For text-to-CAD generation, following prior works~\cite{khan2024text2cad,guan2025cadcoder,xie2025text}, we report \textbf{Invalid Ratio (IR)} for generation reliability and \textbf{Chamfer Distance (CD)} for geometric fidelity. IR counts programs that fail execution, fail CadQuery solid construction, or cannot be converted into a valid evaluation mesh. CD is computed as symmetric squared Chamfer distance between sampled surface point clouds after shape normalization, and is reported in units of $\times 10^3$.

To further evaluate intent-oriented generation quality, we propose \textbf{API Recovery (AR)} for feature recovery. AR is computed as a feature-level F1 score between predicted and ground-truth semantic feature classes after normalizing equivalent CadQuery implementations. For relation preservation, we propose \textbf{Equal Relation Score (ERS)} for equality-like shared-parameter relations and \textbf{Parametric Relation Score (PRS)} for lightweight derived relations.

For zero-shot editing on IntentCAD-Edit-1K, each representation uses its corresponding generation checkpoint without additional editing training. We use \textbf{Edit Success Rate (ES)} as the primary metric, defined as the percentage of edit samples whose outputs are executable and satisfy the target edit specification after variable resolution and normalization of equivalent implementations. We report FE@ES, CE@ES, and DE@ES for the three edit categories introduced in part of \textbf{IntentCAD-Edit-1K Construction}, and retain Median CD only as a weak compatibility metric because global point-cloud distance is often insensitive to small but functionally important local edits. Appendix~B provides the full normalization rules and metric definitions.

\vspace{-1em}
\subsection{Diagnostic Baseline Comparison}
\vspace{-1em}
We first examine how existing text-to-CAD models and general-purpose LLMs transfer to intent-oriented prompts. We evaluate frontier LLMs with direct prompting and include both CAD-Coder~\cite{guan2025cadcoder} and the text-conditioned setting of cadrille~\cite{kolodiazhnyi2025cadrille}, two dedicated models fine-tuned for CadQuery generation on their own datasets. Since these CAD-specific models are trained on different text--code distributions, this comparison is intended as a transfer diagnostic rather than a controlled representation comparison. While both IntentCAD-100K and CAD-Coder's dataset use expert-level text instructions, IntentCAD-100K prompts explicitly describe common geometric features and lightweight relation structures for intent-oriented text-to-CAD generation.

\begin{table}[h]
\centering
\small
\caption{Diagnostic baseline comparison on IntentCAD-100K. CD metrics are $\times 10^3$.}
\label{tab:main_results}
\begin{tabular}{ccccc}
\toprule
Setting & IR.\% $\downarrow$ & Mean CD$\downarrow$ & Median CD$\downarrow$ & AR.\% $\uparrow$ \\
\midrule
GPT-5.4       & 41.85 & 6.21 & 0.13 & 71.97 \\
GPT-5.4-mini      & 50.20 & 38.72 & 0.20 & 69.17 \\
Qwen3.5-397B-A17B   & 41.70 & 12.56 & 0.15 & 75.25 \\
Qwen3.5-plus     & 41.74 & 12.49 & 0.14 & 76.43 \\
\midrule
CAD-Coder~\cite{guan2025cadcoder} & 1.29 & 145.04 & 102.95 & 6.11 \\
cadrille~\cite{kolodiazhnyi2025cadrille} & 24.43 & 208.31 & 116.00 & 0.16 \\
\midrule
Qwen2.5-7B        & 99.98 & 210.04 & 210.04 & 0.00 \\
\textbf{+IntentCAD SFT}  & 0.32 & 0.61 & \textbf{0.12} & 99.94\\
\midrule
 Qwen3.5-9B         & 99.86 & 253.15 & 136.00 & 36.36 \\
\textbf{+IntentCAD SFT}  & \textbf{0.20} & \textbf{0.57} & \textbf{0.12} & \textbf{99.96} \\
\bottomrule
\end{tabular}
\end{table}

Table~\ref{tab:main_results} shows that intent-oriented prompts are not handled reliably by existing systems. General-purpose LLMs suffer from high invalid ratios, indicating that strong general code-generation ability does not directly translate into reliable CadQuery generation under intent-oriented instructions. CAD-Coder and cadrille achieve much lower invalid ratio, but their worse CD and AR suggest limited transfer to our intent-oriented setting, where both geometric fidelity and semantic feature recovery are required. After IntentCAD SFT, both open-source backbones obtain much lower invalid ratios and substantially better geometric and feature recovery.

These results motivate IntentCAD-100K as a challenging intent-oriented benchmark, but they do not by themselves isolate which part of the representation matters. We therefore turn to controlled comparisons in Section~\ref{sec:representation_ablation}, where the data split, backbone, training strategy, and inference setting are fixed, and the supervision representation is varied.

\subsection{Controlled Ablation of Target Representations}
\label{sec:representation_ablation}

\begin{table*}[t]
\centering
% \small
\scriptsize
\setlength{\tabcolsep}{4pt}
\caption{Controlled comparison of representation granularity. Values are point estimates with 95\% bootstrap confidence intervals in brackets where available. CD metrics are $\times 10^3$.}
\label{tab:ablation}
\begin{tabular}{lcccccc}
\toprule
Rep. & IR.\%$\downarrow$ & Mean CD$\downarrow$ & Median CD$\downarrow$ & AR.\%$\uparrow$ & ERS.\%$\uparrow$ & PRS.\%$\uparrow$ \\
\midrule
$-$ & 2.39 [2.09,2.69] & 1.44 [1.12,2.18] & 0.14 & N/A & N/A & N/A \\
$S.$ & 0.22 [0.12,0.33] & \textbf{0.46 [0.28,0.77]} & \textbf{0.12} & \textbf{99.96 [99.93,99.98]} & N/A & N/A \\
$S.+P.$ & \textbf{0.20 [0.12,0.29]} & 0.57 [0.31,0.98] & \textbf{0.12} & 99.96 [99.93,99.98] & 93.35 [91.38,95.15] & 95.25 [94.35,96.14] \\
\bottomrule
\end{tabular}
\end{table*}

Table~\ref{tab:ablation} compares three target representations under the same backbone and fine-tuning recipe: primitive-only programs ($-$), semantic programs ($S.$), and semantic--parametric programs ($S.{+}P.$).
Semantic lifting accounts for the largest improvement in overall generation quality. Compared with primitive-only supervision, $S.$ reduces IR from 2.39\% to 0.22\%, lowers Mean CD from 1.44 to 0.46, and improves Median CD from 0.14 to 0.12. It also achieves 99.96\% API Recovery, indicating that the generated programs preserve the intended semantic feature classes when these features are exposed in the target code.
Parametric lifting mainly affects relation recovery rather than standard geometric metrics. Moving from $S.$ to $S.{+}P.$ keeps IR, AR, and Median CD essentially unchanged. Mean CD is slightly higher for $S.{+}P.$, but the confidence intervals overlap and the median remains the same. In contrast, $S.{+}P.$ achieves 93.35\% ERS and 95.25\% PRS, showing that explicitly represented shared/equal parameters and lightweight derived relations can be recovered during generation. These metrics are not applicable to $-$ or $S.$ because the corresponding relation targets are not exposed in those representations.

Overall, semantic lifting improves reliability and geometric fidelity, while parametric lifting exposes reusable relation structure with little change to the primary generation metrics. This division supports the use of semantic--parametric targets for intent-oriented text-to-CAD generation and motivates the editability analysis in the next section. Appendix~D reports auxiliary syntax validity and completion-token results, which provide additional comparisons across representations.

\subsection{Zero-shot Intent-Preserving Editing}
\label{sec:editability}

We further evaluate whether the learned target representation supports local design refinement without edit-specific training. This experiment is not intended to solve general-purpose CAD editing, unlike prior works that train specialized editors for local infilling or masked-region modification from natural-language instructions~\cite{yuan2025cad,zhang2025geocad,zhang2024flexcad}. Instead, it tests whether a model trained only for text-to-CAD generation can modify an intended feature while preserving the relevant semantic and parametric structure exposed by its representation.

We restrict the edit scope to the code-level intent signals studied in this paper: semantic features, equality-like relations over reusable parameters, and lightweight derived parametric relations. Given a source program and an edit instruction, the model outputs the full edited program using the corresponding generation checkpoint, with no additional editing fine-tuning. Each instruction explicitly refers to the edited object using its loop name in the source program. This intentionally factors out open-ended edit-target localization and isolates whether each representation supports the requested local modification once the target is identified.

\begin{table}[t]
\centering
\scriptsize
\setlength{\tabcolsep}{4pt}
\caption{
Quantitative evaluation on IntentCAD-Edit-1K. Values are point estimates with 95\% bootstrap confidence intervals in brackets where available. CD metrics are $\times 10^3$. FE@ES, CE@ES, and DE@ES report Edit Success Rate on feature, consistency, and derived-relation edits, respectively.
}
\label{tab:edit_overall}
\begin{tabular}{lcccccc}
\toprule
Rep. & IR.\%$\downarrow$ & Median CD$\downarrow$ & ES.\%$\uparrow$ & FE@ES$\uparrow$ & CE@ES$\uparrow$ & DE@ES$\uparrow$ \\
\midrule
$-$ & 36.30 [33.30,39.30] & 0.14 & 41.80 [38.70,44.90] & 48.79 [43.96,53.38] & 39.23 [33.76,44.69] & 34.18 [28.73,40.00] \\
$S.$  & 7.00 [5.50,8.60] & \textbf{0.12} & 88.90 [86.90,90.80] & \textbf{95.17 [93.00,97.10]} & 91.00 [87.78,93.89] & 77.09 [72.00,81.82] \\
$S.{+}P.$  & \textbf{4.70 [3.40,6.00]} & \textbf{0.12} & \textbf{92.50 [90.80,94.10]} & 92.51 [89.86,94.93] & \textbf{95.82 [93.57,97.75]} & \textbf{88.73 [84.73,92.36]} \\
\bottomrule
\end{tabular}
\end{table}

\begin{figure}[t]
    \centering
    \includegraphics[width=1\textwidth]{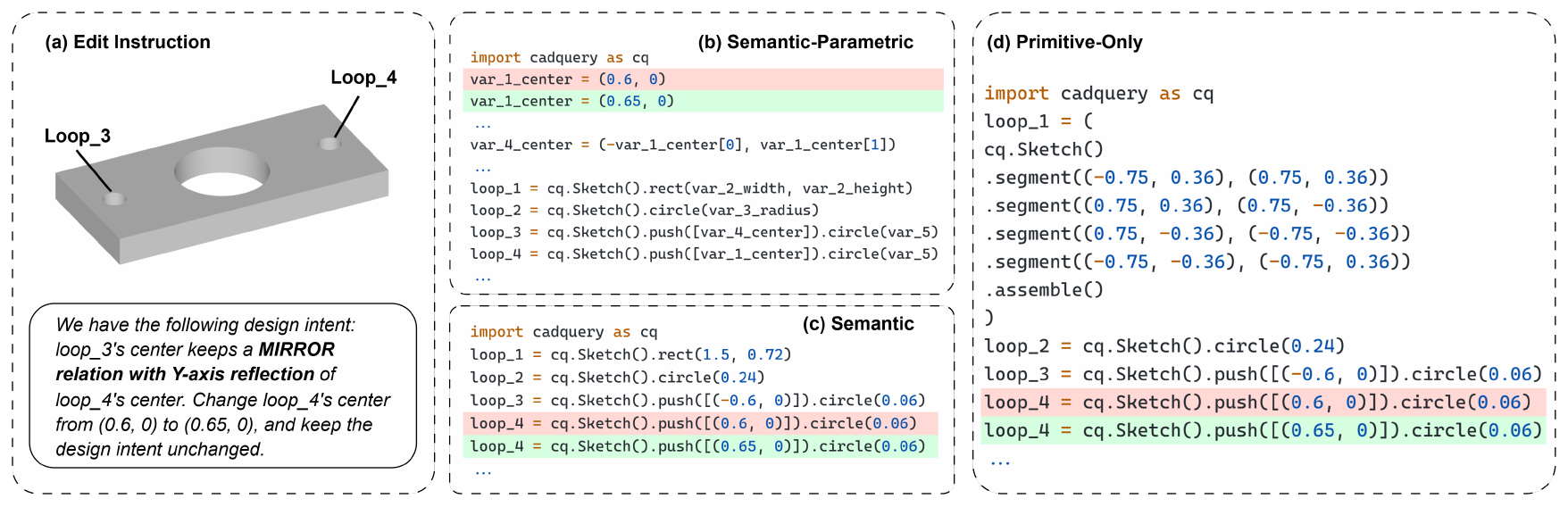}
    \caption{
    Examples of zero-shot intent-preserving edits under different target representations. Semantic targets expose feature-level operations for local modification, while semantic--parametric targets make reusable relations explicit for consistency-preserving refinements.
    }
    \label{fig:edit}
\end{figure}

\paragraph{Results.}
Table~\ref{tab:edit_overall} shows that semantic lifting leads to a large improvement over primitive-only code in zero-shot editing. Overall ES increases from 41.80\% to 88.90\%, while FE@ES increases from 48.79\% to 95.17\%. IR also drops from 36.30\% to 7.00\%, suggesting that semantic feature-level programs provide a more reliable basis for full-program editing than verbose primitive traces. Paired bootstrap intervals in Appendix~D support the improvement across overall ES and the three edit categories.
Parametric lifting provides a more targeted benefit. Moving from $S.$ to $S.{+}P.$ keeps the geometry proxy comparable, with the same Median CD of 0.12, and FE@ES remains similar. The main gains appear in relation-aware edits: CE@ES increases from 91.00\% to 95.82\%, and DE@ES increases from 77.09\% to 88.73\%, with paired bootstrap confidence intervals excluding zero. These results suggest that explicit parameter and relation structure is useful when local refinements must preserve equality-like or derived relations. More qualitative results refer to Appendix~\ref{app:additional_qualitative}.

These results suggest that representation granularity affects zero-shot intent-preserving editing under our controlled benchmark. Thus, editing serves as a downstream validation of the representation: semantic structure makes local feature edits more reliable, and explicit parametric structure helps preserve relations during refinement. A finer breakdown by edit subtype is provided in Appendix~D.

\subsection{Visually Grounded Editing}
\label{sec:vis_grounded}

To isolate editing performance from target localization, the preceding zero-shot experiment uses explicit loop identifiers. We extend this evaluation to visually grounded editing using 200 cases from IntentCAD-Edit-1K with at least two candidate features (80 FE, 60 CE, and 60 DE). Loop identifiers are replaced with spatial descriptions while preserving the edit operations and target values. GPT-5.4 edits each aligned source program using the same four-view rendering and instruction. We also use GPT-5.4 to separately rate source program editability on a 0--10 scale without representation labels. Localization checks whether the intended target loop is modified, while ES additionally requires executable and correct edits, including relation--dependent updates.

\begin{table}[htbp]
\centering
\small
\setlength{\tabcolsep}{4pt}
\caption{Visually grounded editing on 200 cases.
VLM scores range from 0 to 10.}
\label{tab:vlm_overall}
\begin{tabular}{lrrrrrr}
\toprule
Rep. & Loc. (\%) $\uparrow$ & ES (\%) $\uparrow$ &
FE@ES (\%) $\uparrow$ & CE@ES (\%) $\uparrow$ &
DE@ES (\%) $\uparrow$ & VLM $\uparrow$ \\
\midrule
$-$ & 69.50 & 36.50 & 33.75 & 33.33 & 43.33 & 6.58 \\
$S.$ & 84.00 & 66.00 & \textbf{86.25} & 48.33 & 56.67 & 7.86 \\
$S.{+}P.$ & \textbf{91.50} & \textbf{78.50} & 78.75 &
\textbf{73.33} & \textbf{83.33} & \textbf{8.94} \\
\bottomrule
\end{tabular}
\end{table}

\begin{figure}[t]
\centering

{\small
\textbf{Edit instruction:}
\emph{Set the \textbf{height} of the outer rectangular profile on the top face to 0.942856, while preserving the equal-dimension relations among the related features. Leave all unrelated geometry unchanged.}
\par}

\vspace{0.3em}

\begin{minipage}[t]{0.235\linewidth}
    \centering
    \includegraphics[width=0.82\linewidth]{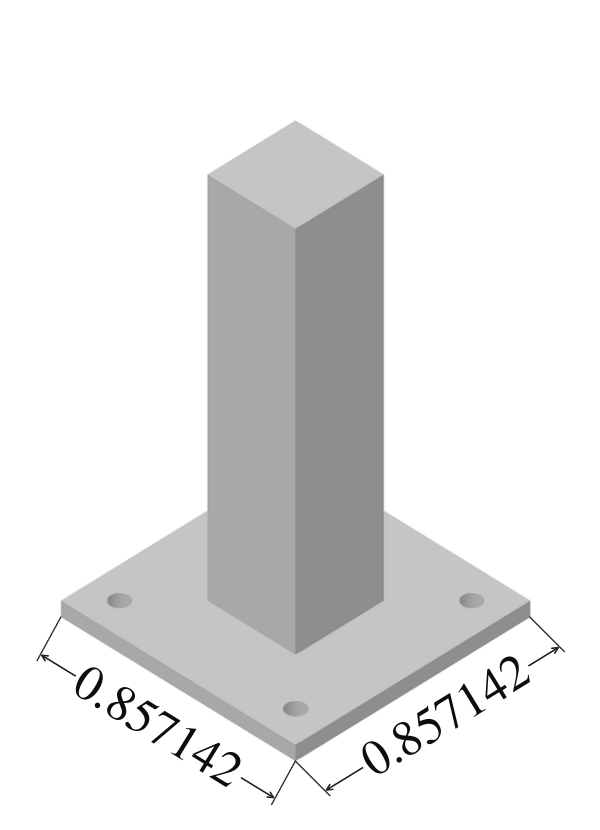}

    \vspace{-0.3em}
    {\scriptsize
    $\leftrightarrow$ \texttt{EQUAL(width,height)}
    \par}
    \vspace{0.2em}
    {\small (a) Source\par}
\end{minipage}
\hfill
\begin{minipage}[t]{0.235\linewidth}
    \centering
    \includegraphics[width=0.82\linewidth]{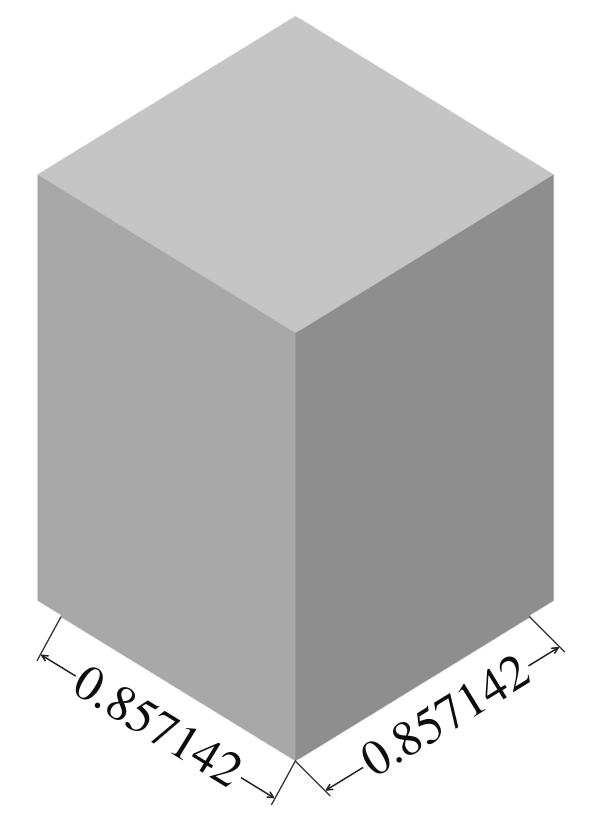}

    \vspace{-0.3em}
    {\scriptsize
    \textcolor{red!80!black}{{$\times$}
    Shape Mismatch}
    \par}
    \vspace{0.2em}
    {\small (b) $-$\par}
\end{minipage}
\hfill
\begin{minipage}[t]{0.235\linewidth}
    \centering
    \includegraphics[width=0.82\linewidth]{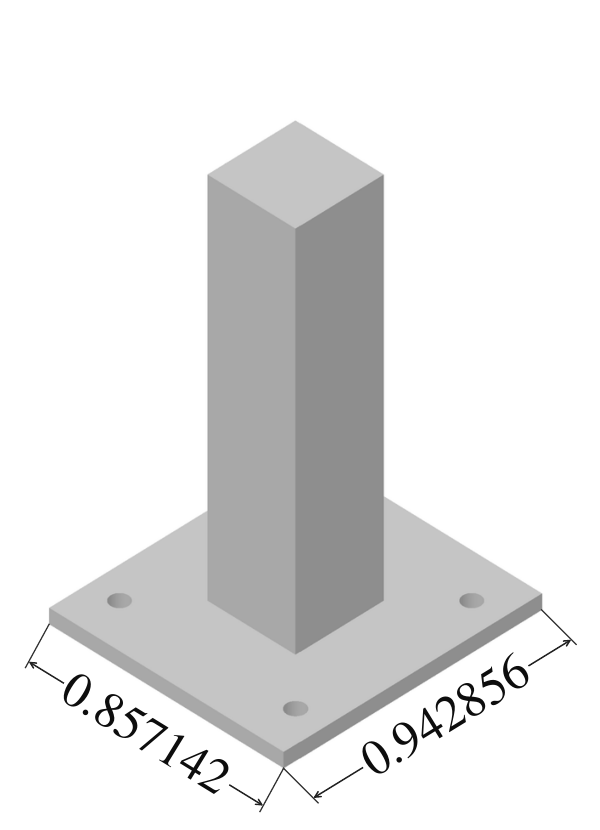}

    \vspace{-0.3em}
    {\scriptsize
    \textcolor{yellow!70!black}{{\textbf{!}}
    Relation Mismatch}
    \par}
    \vspace{0.2em}
    {\small (c) $S.$\par}
\end{minipage}
\hfill
\begin{minipage}[t]{0.235\linewidth}
    \centering
    \includegraphics[width=0.82\linewidth]{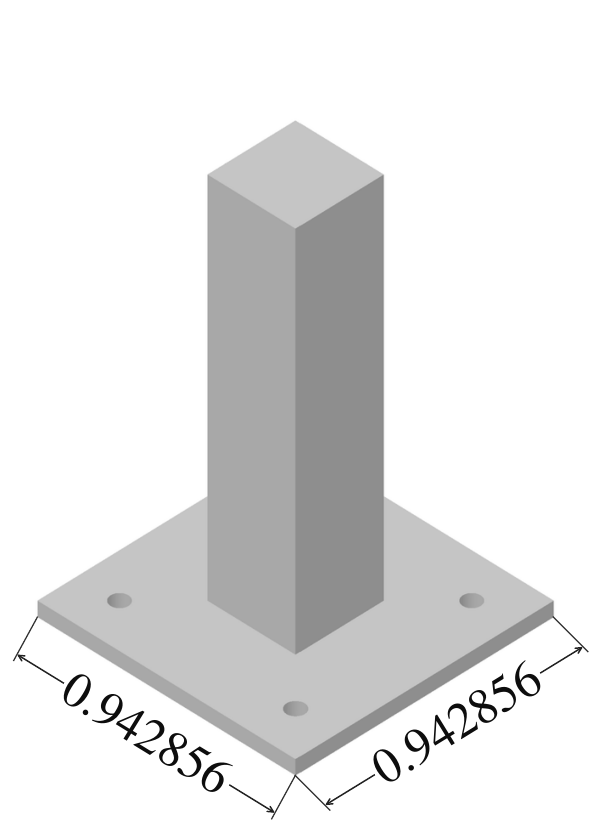}

    \vspace{-0.3em}
    {\scriptsize
    \textcolor{green!80!black}{{$\checkmark$}
    Success}
    \par}
    \vspace{0.2em}
    {\small (d) $S.{+}P.$\par}
\end{minipage}

\caption{Qualitative comparison of visually grounded editing
across different source representations.}
\label{fig:vlm_localization}
\end{figure}

Table~\ref{tab:vlm_overall} shows that semantic lifting improves localization from 69.50\% to 84.00\% and ES from 36.50\% to 66.00\%, while parametric lifting further raises them to 91.50\% and 78.50\%, respectively.
The additional ES gains mainly occur in CE and DE, although $S.$ performs better on FE. Higher VLM ratings also suggest improved perceived editability. Figure~\ref{fig:vlm_localization} illustrates how explicit parameter relations help preserve dimensional consistency during editing.

\section{Limitations}
\label{sec:limitation}

Several limitations define the current scope of Intent2CAD. 
First, it supports only a conservative set of commonly recoverable semantic templates, leaving many long-tail engineering features in primitive form. 
Extending semantic coverage to richer feature spaces remains important future work. 
Second, the text descriptions are primarily expert-level prompts, and generalization to casual or underspecified instructions remains unstudied. 
Third, our use of \emph{intent} is limited to recoverable code-level signals rather than full design history or constraint-solving processes. 
Finally, the conservative pipeline reduces incorrect supervision, but may leave some valid relations unlifted.

\section{Conclusion}
\label{sec:conclusion}

This paper studied how semantic--parametric representation affects text-to-CAD generation under a controlled fine-tuning setup. 
Intent2CAD changes the supervision target rather than the model architecture or decoding procedure, representing CAD programs as executable CadQuery code with recoverable semantic features, reusable parameters, lightweight parametric relations, and low-level fallbacks. 
Our experiments suggest that semantic lifting improves reliability, geometric fidelity, feature recovery, and feature-level editing, while parametric lifting mainly benefits relation recovery and relation-aware edits. 
These findings provide evidence that the symbolic form of CAD supervision can affect the structure learned by text-to-CAD models. 
Within the scope of recoverable code-level intent signals, structured supervision appears to be a promising direction for generating CAD programs that are not only executable, but also more amenable to later refinement.

\bibliographystyle{unsrtnat}

\section{Acknowledgments and Disclosure of Funding}
This research was supported by the National Key R\&D Program of China (No. 2023YFB4604800).
The authors declare no competing interests.

\bibliography{refs}

\appendix
\label{appendix}

\section{Additional Dataset and Pipeline Details}
\label{app:dataset}

\subsection{Additional IntentCAD-100K Statistics}
\label{app:dataset_statistics}

\paragraph{Data Distribution.}
Table~\ref{tab:dataset_summary} shows that the three sets exhibit highly consistent statistics, including the proportions of semantic and parametric lifted samples. Table~\ref{tab:loop_type_dist} further reports the loop-type distribution of the three sets.

\begin{table}[h]
\centering
\renewcommand{\arraystretch}{1.05}

\begin{minipage}[h]{0.4\linewidth}
\centering
% \small
% \setlength{\tabcolsep}{4pt}
% Parametric lifting is applied only when semantic lifting provides reliable editable variables; 
\caption{Distribution of instances with semantic lifting (S.), parametric lifting (P.), and both semantic and parametric lifting (S.+P.) in each split (\%).}
\label{tab:dataset_summary}
\begin{tabular}{lccc}
\toprule
% \textbf{Split} & 
% \begin{tabular}{@{}c@{}}\textbf{Semantic} \\ \textbf{Lifted}\end{tabular} & 
% % \begin{tabular}{@{}c@{}}\textbf{Fully} \\ \textbf{Lifted}\end{tabular} & 
% \begin{tabular}{@{}c@{}}\textbf{Param.} \\ \textbf{Relation}\end{tabular} & 
\textbf{Set} & \textbf{S.} & \textbf{P.} & \textbf{S.+P.} \\

% \textbf{Both} \\
\midrule
Train.      & 69.65 & 25.97 & 18.69 \\
Val.        & 70.60 & 26.14 & 19.24 \\
Test.       & 69.44 & 25.93 & 18.47 \\
\bottomrule
\end{tabular}
\end{minipage}\hfill
\begin{minipage}[h]{0.58\linewidth}
\centering
\small
\setlength{\tabcolsep}{2pt}
\caption{Loop type distribution across dataset (\%). Fall. denotes fallback loops with segments and arcs. Poly. and Reg. denotes irregular and regular polygons, respectively.}
\label{tab:loop_type_dist}
\begin{tabular}{lcccccccc}
\toprule
 & \multicolumn{2}{c}{\textbf{Low-level}} & \multicolumn{6}{c}{\textbf{High-level}} \\
\cmidrule(lr){2-3} \cmidrule(lr){4-9}
\textbf{Set} & \textbf{Circle} & \textbf{Fall.} & \textbf{Rect.} & \textbf{Poly.} & \textbf{Fillet} & \textbf{Reg.} & \textbf{Array} & \textbf{Slot} \\
\midrule
Train.      & 37.34 & 14.51 & 32.63 & 11.80 & 1.48 & 0.99 & 0.69 & 0.55 \\
Val.        & 36.81 & 14.76 & 33.16 & 11.40 & 1.66 & 0.90 & 0.76 & 0.55 \\
Test.       & 37.90 & 14.22 & 32.70 & 11.23 & 1.51 & 1.10 & 0.75 & 0.60 \\
\bottomrule
\end{tabular}
\end{minipage}
\end{table}

\paragraph{Semantic lifting coverage.}
Table~\ref{tab:semantic_coverage} reports sample-level semantic lifting coverage, complementing the loop-level distribution in Table~\ref{tab:loop_type_dist}.

\begin{table}[h]
\centering
\small
\setlength{\tabcolsep}{4pt}
\caption{Sample-level semantic lifting coverage.}
\label{tab:semantic_coverage}
\begin{tabular}{lcccc}
\toprule
Set & Avg. semantic features & Avg. circle loops & Avg. fallback loops & Fully fallback samples \\
\midrule
Train & 1.12 & 0.99 & 0.34 & 5,070 (6.29\%) \\
Val. & 1.13 & 0.99 & 0.35 & 622 (6.17\%) \\
Test & 1.12 & 1.00 & 0.33 & 615 (6.10\%) \\
\bottomrule
\end{tabular}
\end{table}

\paragraph{Source constraint distribution.}
Table~\ref{tab:constraint_type_distribution} summarizes the source CAD constraints extracted before lifting, grouped by their primary use in our pipeline. Geometric constraints provide evidence for local feature lifting, while dimensional and parameter-related constraints support relation proposal and parameter rewriting. These source constraints are not one-to-one labels of
the final explicit relations in code.

\begin{table}[h]
\centering
\small
\setlength{\tabcolsep}{5pt}
\caption{
Aggregate distribution of source CAD constraints before lifting, grouped into
geometric and dimensional sketch constraints.
}
\label{tab:constraint_type_distribution}
\begin{tabular}{lrr@{\hspace{1.2em}}lrr}
\toprule
\multicolumn{3}{c}{Geometric constraints} &
\multicolumn{3}{c}{Dimensional constraints} \\
\cmidrule(lr){1-3}\cmidrule(lr){4-6}
Type & Count & Overall (\%) &
Type & Count & Overall (\%) \\
\midrule
PARALLEL      & 155,431 & 17.36 & LENGTH   & 147,184 & 16.44 \\
HORIZONTAL    & 125,251 & 13.99 & DISTANCE & 83,202  & 9.29  \\
PERPENDICULAR & 93,149  & 10.40 & DIAMETER & 77,295  & 8.63  \\
TANGENT       & 68,346  & 7.63  & RADIUS   & 21,991  & 2.46  \\
VERTICAL      & 47,689  & 5.33  & ANGLE    & 5,477   & 0.61  \\
CONCENTRIC    & 26,617  & 2.97  &          &         &       \\
EQUAL         & 22,428  & 2.51  &          &         &       \\
MIRROR        & 21,249  & 2.37  &          &         &       \\
\midrule
Total         & 560,160 & 62.56 & Total    & 335,149 & 37.43 \\
\bottomrule
\end{tabular}
\end{table}

\subsection{Duplicate Removal}
\label{app:deduplication}

Duplicate removal is performed after primitive CadQuery reconstruction and validity filtering, but before lifting and data splitting, ensuring that all three supervision representations remain aligned. Each primitive program is converted into an ordered construction signature from its operations and geometric arguments; floating-point values are rounded to four decimals and the serialized signature is hashed with SHA256. Duplicate hashes are removed after deterministic sorting.

This is construction-level rather than source-code or full geometric-equivalence deduplication: formatting differences are ignored, but programs with different operation orders or construction histories may remain distinct. This choice matches our setting, where CAD supervision depends on both final geometry and construction parameterization.

\subsection{Semantic–Parametric Lifting Details}

\paragraph{Supported semantic features.}
Semantic lifting supports rectangles, regular and irregular polygons, slots, fillets, and arrays. Unsupported or ambiguous primitive groups are preserved as low-level fallbacks.

\paragraph{Supported parametric relations.}
Parametric lifting targets EQUAL, CONCENTRIC, MIRROR, DISTANCE, and RATIO relations. These relations are represented in code as shared variables or simple derived expressions after validation. Implementation details of rewrite proposal and validation are described below.

\paragraph{LLM-Assisted Local Rewrite Prompt.}
\label{app:llm_prompt}

Stage II uses an LLM only as a constrained local rewrite proposer. 
The LLM receives the Stage-I program and primitive-level constraints, and returns a structured rewrite plan over variable definitions. 
The system prompt is shown in Prompt~\ref{prompt:stage2_system}. 
The user message contains the Stage-I CadQuery program and the corresponding serialized constraints.

\begin{promptbox}{Prompt~\refstepcounter{prompt}\theprompt: Prompt for Stage-II relation induction}
\label{prompt:stage2_system}
\small

\underline{\textit{\textbf{User message template:}}}

\medskip
\noindent\textbf{CadQuery Code.}

\smallskip
\noindent\texttt{\{code\_text\}}

\medskip
\noindent\textbf{Constraints.}

\smallskip
\noindent\texttt{\{constraints\_summary\_json\}}

\medskip
\noindent Identify all valid variable substitutions based on \texttt{EQUAL}, \texttt{CONCENTRIC}, \texttt{DISTANCE}, \texttt{LENGTH}, \texttt{RADIUS}, \texttt{DIAMETER}, or \texttt{MIRROR}. 
Do not infer directly from \texttt{PARALLEL}, \texttt{PERPENDICULAR}, \texttt{TANGENT}, or other non-substitution constraints. 
Output only the JSON object.

\medskip
\hrule
\medskip

\underline{\textit{\textbf{System prompt:}}}
\medskip

\textbf{Role.}
You are a CAD parametrization expert. Analyze an already high-level CadQuery sketch script together with sketch-level geometric constraints, and rewrite only variable definitions into shared variables or simple expressions.

\medskip
\textbf{Task.}
The code already uses high-level APIs such as \texttt{rect}, \texttt{circle}, \texttt{slot}, \texttt{regularPolygon}, and \texttt{fillet}. 
Do not change code structure, API calls, loops, or body logic. 
Only improve parameter definitions \texttt{var\_*}. 
The goal is to recover human-intuitive parametric design intent, not to numerically fit arbitrary relations between variable values.

\medskip
\textbf{Input notes.}
Code comments map constraint tags to loops, e.g., \texttt{\# 'JGV', 'JGp' -> loop\_2}. 
Use these comments to connect constraint entities to the corresponding high-level geometry. 
A constraint entity may correspond to multiple sketch tags because one geometric entity can be represented by multiple sketch segments. 
Treat such tags as one grouped entity.

\medskip
\textbf{Rewrite rules.}
Only modify \texttt{var\_*} definitions. 
Do not change geometry, APIs, loops, or body logic. 
Infer parameter meaning from the existing high-level API, variable names, and provided sketch constraints. 
Preserve geometry exactly after substitutions, with tolerance $\leq 10^{-5}$. 
Prefer shared parameters or simple human-like expressions over repeated constants. 
Do not replace expressions entirely with bare numbers. 
If evidence is weak or ambiguous, make no change.

\medskip
\textbf{Constraint use.}
\texttt{EQUAL} is strong evidence for shared parameters. 
\texttt{CONCENTRIC}, \texttt{DISTANCE}, and \texttt{MIRROR} may justify shared centers or simple symmetric expressions. 
\texttt{LENGTH}, \texttt{RADIUS}, and \texttt{DIAMETER} are dimension evidence only; use them only for equality or clearly intended simple relations. 
Other constraints are supporting context only, not sole evidence.

\medskip
\textbf{Compatibility.}
Relations may be inferred between geometrically compatible quantities, not only identical property names, e.g., width--height, radius--radius, radius--diameter, line length--width/height, and distance--width/height/diameter. 
Do not relate incompatible quantities such as size--angle or size--side count.

\medskip
\textbf{Output format.}
Return a JSON object with one required field, \texttt{changes}. 
If no changes are justified, return an empty list.

\begin{verbatim}
{
  "changes": [
    {
      "variable": "<variable_name>",
      "new_value": "<expression_or_variable>",
      "relation_type": "shared" | "expression",
      "primary_constraints": [
        {
          "type": "EQUAL" | "CONCENTRIC" | "DISTANCE" |
                  "RADIUS" | "LENGTH" | "DIAMETER" | "MIRROR",
          "entities": ["<tag1>", "<tag2>"]
        }
      ],
      "reason": "<brief explanation under 20 words>"
    }
  ]
}
\end{verbatim}

\end{promptbox}

\subsection{Validation and Text Description Generation Rules}
\label{app:validation}

\paragraph{Execution check.}
We execute each candidate CadQuery program. A program passes execution if it produces a valid non-empty solid whose volume can be queried successfully within 30 seconds.

\paragraph{Geometric consistency.}
Since Chamfer Distance is randomly sampled, we compare the candidate program $\tilde{p}$ against the Stage-I program $p^{(1)}$ using deterministic values, such as solid volume, the six axis-aligned bounding-box coordinates, and the three center-of-mass coordinates. A rewrite is accepted only if every checked scalar invariant differs under $10^{-5}$ tolerance.

\paragraph{Locality check.}
We parse both $p^{(1)}$ and $\tilde{p}$ into ASTs and verify that the geometric operation sequence, loop identifiers, feature calls, extrusion operations, Boolean operations, and sketch topology remain unchanged. The rewrite may only introduce or modify variable assignments and parameter expressions. It is rejected if it adds, removes, reorders, or changes geometric operations.

\paragraph{Text Description Generation.}
\label{app:text_generation}

For each accepted program, we use GPT-5.4 to generate one intent-oriented engineering description from the accepted CadQuery script and the recovered constraint summary. The generator is instructed to verbalize only code-supported geometry, dimensions, construction order, body operations, and explicitly represented relations. It is also instructed not to mention Python, CadQuery, API names, variable names, loop identifiers, or other code-level symbols. Numeric-equality-only groups are not described as shared relations unless they are explicitly supported by the accepted program or constraint summary.

\subsection{Stage-wise Coverage and Validation}
\label{app:pipeline_coverage}

Stage I fully recognizes the semantic structure in 82.68\% of the retained IntentCAD-100K programs. Incomplete recognition arises from unsupported feature patterns, insufficient constraint evidence, or geometric deviations beyond the matching tolerance. Unrecognized portions remain in primitive form rather than being assigned uncertain semantic labels.

Stage II proposes at least one relation rewrite for 34.24\% of Stage-I programs. The absence of a proposal may reflect insufficient evidence or relations already encoded by semantic features. For example, parallelism and perpendicularity are intrinsic to \texttt{rect}. This percentage measures proposal coverage rather than the proportion of programs with accepted parametric lifting.

Among 78,930 candidate rewrites, 70,238 (88.99\%) pass deterministic validation. The remaining candidates are rejected when substitutions violate executability, geometric consistency, or locality. Rejected rewrites retain the Stage-I representation, while unsuccessful semantic lifting preserves the corresponding primitive operations. These fallbacks are distinct from the initial filtering of raw sequences that cannot be converted into executable programs.

\section{Details of Metric Definitions}

\subsection{Generation Metrics}
\label{app:generation_metrics}

\paragraph{Execution Validity.}
Invalid Ratio (IR) is the fraction of predicted programs that fail to execute into a valid CadQuery solid. A prediction is considered executable if running the program produces an object whose solid volume can be queried successfully.

\paragraph{Chamfer Distance.}
For executable predictions, we tessellate both the predicted and ground-truth CadQuery solids with a tolerance of $0.001$, uniformly sample 8192 surface points, normalize each point cloud by its maximum bounding-box extent, and compute symmetric squared Chamfer Distance. Following prior text-to-CAD works, reported CD values are multiplied by $10^3$.

\paragraph{API Recovery.}
API Recovery (AR) measures whether the generated code recovers the semantic feature calls  used by the ground-truth program. We parse the predicted and target programs into Python ASTs and extract multisets of canonical feature labels. Equivalent API interfaces are mapped to the same feature label when applicable, e.g., \texttt{slot} and \texttt{slot2D} are both mapped to \texttt{slot}, \texttt{pushPoints}, \texttt{rarray}, and \texttt{polarArray} are mapped to \texttt{array}. Given ground-truth and predicted feature multisets $M_g$ and $M_p$, AR is computed as the F1 score:
\[
P=\frac{|M_g\cap M_p|}{|M_p|},\quad
R=\frac{|M_g\cap M_p|}{|M_g|},\quad
AR=\frac{2PR}{P+R}.
\]
Samples without ground-truth canonical features are excluded from AR calculation.

\paragraph{Equal Relation Score.}
Equal Relation Score (ERS) measures preservation of explicit shared-parameter structure. We identify each variable that is referenced by two or more loop definitions as a shared parameter. Each shared parameter is represented by a structural signature consisting of its definition skeleton, its API argument roles, and the number of loops that reference it. Concrete variable names and numeric constants are abstracted in the skeleton, so the score focuses on whether the reusable parameter structure is preserved rather than whether the exact identifier names match. ERS is the fraction of ground-truth shared-parameter signatures greedily matched by the prediction. ERS is only reported for samples whose ground-truth programs contain shared parameters.

\paragraph{Parametric Relation Score.}
Parametric Relation Score (PRS) measures preservation of derived parameter expressions. A derived parameter is a variable whose right-hand side references at least one other variable. We represent each derived parameter by its expression skeleton and the number of referenced variables, and compute PRS as the fraction of ground-truth derived-relation signatures matched by the prediction. PRS is only reported for samples whose ground-truth programs contain derived parameter relations.

\subsection{Edit Metrics}
\label{app:edit_metrics}

\paragraph{Edit Success Rate.}
Edit Success Rate (ES) measures whether the generated program correctly realizes the requested edit. A prediction is counted as successful if it satisfies two conditions: (1) the program executes into a valid CAD solid; and (2) the loops affected by the edit match the deterministic target program. The affected loops include the directly edited loop and any loops that should change to preserve the declared relation.

The match is not based on string matching. For semantic loops, we compare the recovered feature type and resolved feature attributes, such as center, width, height, radius, angle, polygon points, slot dimensions, and fillet radius; variable names and expression surface forms are ignored after values are resolved. For primitive loops, we compare the resulting low-level segment/arc construction, and treat cyclic shifts and reversed traversal of the same closed loop as equivalent. Numeric values are compared with a tolerance of $10^{-5}$. The final ES is the percentage of edit samples that satisfy these conditions.

\section{Edit Benchmark Details}

\subsection{Edit Benchmark Construction Pipeline}
% triples, selected_ops.tsv fields, validation.
\label{app:edit_benchmark}

We construct the edit benchmark by deterministic synthesis from the subset of IntentCAD-100K test set that contains both semantic and parametric lifting, without using LLM-generated targets. The purpose is to obtain aligned edit triples where the same edit instruction is applied to primitive-only, semantic-only, and semantic--parametric representations.

The synthesis follows a conservative pipeline similar to Stage I. We first identify editable local semantic features through AST parsing, and use loop/tag alignment to trace the affected loops across the three representations. The edit therefore is applied at the semantic level and then projected to the other representations: $S.{+}P.$ updates shared or derived variable assignments when applicable, $S.$ rewrites the corresponding feature parameters, and the primitive representation ($-$) regenerates the involved tagged loops as equivalent low-level primitive chains.

We keep only candidates that pass several validations: all source and target programs must be syntactically valid and executable, the target must introduce a non-trivial geometric change from the source, and the three target programs must remain mutually consistent under a strict geometric tolerance $10^{-5}$.

\subsection{Edit Type Distribution}
Table~\ref{tab:edit_distribution} summarizes the three edit categories. The benchmark is designed to cover local feature changes, equality-style consistency preservation, and derived relation preservation. Table~\ref{tab:edit_subtypes} further reports subtype statistics. Feature edits are approximately balanced between parameter modification and feature replacement, while relation-sensitive edits mainly cover \texttt{EQUAL}, \texttt{CONCENTRIC}, \texttt{RATIO}, \texttt{MIRROR}, and \texttt{DISTANCE} relations.

\begin{table}[h]
\centering
\caption{Distribution of edit types.}
\label{tab:edit_distribution}
\begin{tabular}{lrr}
\toprule
Edit type & Count & Percentage \\
\midrule
Feature Edit (FE) & 414 & 41.40\% \\
Consistency Edit (CE) & 311 & 31.10\% \\
Derived-Relation Edit (DE) & 275 & 27.50\% \\
\bottomrule
\end{tabular}
\end{table}

\begin{table}[h]
\centering
\caption{Subtype distribution of edits. Percentages are computed within each major edit type.}
\label{tab:edit_subtypes}
\begin{tabular}{llrr}
\toprule
Type & Subtype & Count & Within-type \\
\midrule
\multirow{2}{*}{Feature} 
& Parameter modification & 200 & 48.31\% \\
& Feature replacement & 214 & 51.69\% \\
\midrule
\multirow{6}{*}{Consistency}
& \texttt{EQUAL} (height+width) & 74 & 23.79\% \\
& \texttt{EQUAL} (height) & 63 & 20.26\% \\
& \texttt{EQUAL} (radius) & 63 & 20.26\% \\
& \texttt{EQUAL} (width) & 62 & 19.94\% \\
& \texttt{CONCENTRIC} (center) & 37 & 11.90\% \\
& \texttt{EQUAL} (fillet radius) & 8 & 2.57\% \\
& \texttt{EQUAL} (other combinations) & 4 & 1.28\% \\
\midrule
\multirow{4}{*}{Derived-Relation}
& \texttt{RATIO} & 131 & 47.64\% \\
& \texttt{MIRROR} & 128 & 46.54\% \\
& \texttt{DISTANCE} & 16 & 5.82\% \\
\bottomrule
\end{tabular}
\end{table}

\subsection{Edit Instruction Templates}

We use three types of edit template. Each instruction describes the requested modification and explicitly asks the model to keep the remaining relations unchanged.

\paragraph{Feature Edit.}
For parameter modifications, the instruction specifies the edited feature, loop identifier, attribute, old value, and new value: "Change the \{attribute\} of the \{feature\} in \{loop\} from \{old\_value\} to \{new\_value\}, and keep the other supported structures unchanged."

For feature replacements, the instruction specifies the target feature and its required attributes: "Replace the geometry in \{loop\} with a \{new\_feature\} feature, where \{attribute descriptions\}, and keep the design intent unchanged."

\paragraph{Consistency Edit.}
These instructions explicitly state an equality-style relation, such as EQUAL or CONCENTRIC, and ask the model to update one parameter while preserving the relation: "The following relation should be preserved: \{affected slots\} stay \{relation\}. Change \{edited slot\} from \{old\_value\} to \{new\_value\}, and keep the design intent unchanged."

\paragraph{Derived-Relation Edit.}
These instructions specify a derived relation such as MIRROR, RATIO, or DISTANCE, and ask the model to update the source parameter while preserving the relation: "The following relation should be preserved: \{dependent slot\} keeps a \{relation\} relation with \{source slot\}. Change \{source slot\} from \{old\_value\} to \{new\_value\}, and keep the design intent unchanged."

\section{Additional Experiment Result Details}

\subsection{Additional Quantitative Results}
\label{app:additional_quant_results}

Table~\ref{tab:aux_sv_ct} reports some auxiliary quantitative results.  
We include \textbf{Syntax Validity (SV)} to measure Python syntax correctness, and \textbf{Completion Tokens (CT)} as an auxiliary measure of decoding cost during inference. 
These metrics complement the main evaluation, but they are not used as primary indicators of geometric fidelity, semantic feature recovery, relation preservation, or edit success.

\begin{table*}[t]
\centering
\caption{
Auxiliary syntax-validity and decoding-cost results. 
SV reports Python syntax validity, and Mean CT reports average completion tokens.
}
\label{tab:aux_sv_ct}
\begin{tabular}{llcc}
\toprule
Experiment & Setting / Rep. & SV.\%$\uparrow$ & Mean CT$\downarrow$ \\
\midrule
\multicolumn{4}{l}{\textit{Diagnostic Baseline Comparison}} \\
& GPT-5.4       & 99.97 & 341.08 \\
& GPT-5.4-mini      & 99.98 & \textbf{258.74} \\
& Qwen3.5-397B-A17B   & 99.60 & 2109.67 \\
& Qwen3.5-plus     & 99.61 & 2115.25 \\
& CAD-Coder~\cite{guan2025cadcoder} & 99.66 & 270.51 \\
& Qwen2.5-7B        & 99.75 & 381.64 \\
& \textbf{+IntentCAD SFT}  & 99.98 & 371.15 \\
& Qwen3.5-9B         & 91.64 & 1056.28 \\
& \textbf{+IntentCAD SFT}  & \textbf{99.99} & 365.23 \\

\midrule
\multicolumn{4}{l}{\textit{Controlled Ablation of Target Representations}} \\
& $-$ & 97.95 & 713.74 \\
& $S.$ & \textbf{99.99} & \textbf{341.21} \\
& $S.+P.$ & \textbf{99.99} & 365.23 \\

\midrule
\multicolumn{4}{l}{\textit{Zero-shot Intent-Preserving Editing}} \\
& $-$ & 66.20 & 2590.3 \\
& $S.$  & 94.20 & 1334.7 \\
& $S.+P.$  & \textbf{98.50 } & \textbf{1248.5} \\
\bottomrule
\end{tabular}
\end{table*}

For the controlled representation comparison, semantic lifting also reduces decoding cost. 
Mean CT decreases from 713.74 for the primitive-only representation ($-$) to 341.21 for the semantic representation ($S.$). 
For reference, the mean extracted generated-code lengths are 490.85, 339.21, and 363.23 tokens for $-$, $S.$, and $S.+P.$, respectively. 
This suggests that the CT reduction from $-$ to $S.$ is not explained by final code length alone, and is also associated with the more compact and structured generation target introduced by semantic lifting.

Adding parametric lifting slightly increases Mean CT from 341.21 to 365.23. 
This is expected because $S.+P.$ introduces explicit variables and lightweight relations into the target code. 
The same trend appears in the extracted code length, which increases from 339.21 tokens for $S.$ to 363.23 tokens for $S.+P.$.

On the edit benchmark, SV improves from 66.20\% for the primitive-only representation to 94.20\% for $S.$ and 98.50\% for $S.+P.$, indicating that lifted representations also make zero-shot editing less prone to syntactically invalid outputs. 
Mean CT also decreases from 2590.3 to 1334.7 and 1248.5, suggesting that the lifted representations lead to shorter edit responses.

\subsection{Implementation Details.}
All fine-tuning experiments are implemented with a standard LoRA-based supervised fine-tuning recipe~\cite{hu2022lora}. 
Unless otherwise specified, we use Qwen3.5-9B as the base model. 
The backbone parameters are frozen, and only the LoRA adapters are updated. 
LoRA is applied to both attention and feed-forward projection modules, including the query, key, value, output, gate, up, and down projections. 
For each supervision representation, we train a separate LoRA adapter for three epochs using the same optimizer configuration, effective batch construction, learning-rate schedule, and sequence-length setting. 
Thus, the controlled variants differ only in the supervision target representation. 
During inference, all variants use the same decoding configuration and maximum generation length 4096. 
All fine-tuning and evaluation experiments are conducted on 2 NVIDIA A100 40GB GPUs.

\subsection{Bootstrap and Paired Difference Analysis}

All confidence intervals use percentile bootstrap with 10,000 resamples over test examples. For paired representation comparisons, the same example indices are resampled across representations. Positive differences indicate that the main representation performs better: for lower-is-better metrics such as IR and CD, we compute baseline minus main; for higher-is-better metrics, we compute main minus baseline.

For controlled generation, Table~\ref{tab:paired_generation_stats} analyzes representation differences using paired bootstrap confidence intervals. 
Semantic lifting yields supported improvements over primitive-only supervision: the intervals for IR and Mean CD are strictly positive. 
Its role is instead evaluated through relation-oriented metrics: ERS and PRS are only meaningful when explicit shared parameters or derived relations exist, and are therefore reported as absolute scores for $S.+P.$ in Table~\ref{tab:ablation}.

\begin{table}[h]
\centering
\setlength{\tabcolsep}{4pt}
\caption{Paired bootstrap difference confidence intervals for controlled generation comparisons. Positive values indicate improvement of the main representation. CD metrics are $\times 10^3$.}
\label{tab:paired_generation_stats}
\begin{tabular}{llcc}
\toprule
Comparison & Metric & Difference & 95\% CI \\
\midrule
$S.$ vs. $-$ & IR.\% & +2.17 & [1.87, 2.45] \\
$S.$ vs. $-$ & Mean CD & +0.98 & [0.38, 1.62] \\
\midrule
% REVIEW: Author to check the paired IR CI below. The original interval is
% retained in blue at the author's request; the main-text IRs imply +0.02.
$S.{+}P.$ vs. $S.$ & IR.\% & +0.02 & [-0.07, 0.05] \\
$S.{+}P.$ vs. $S.$ & Mean CD & -0.11 & [-0.42, 0.18] \\
$S.{+}P.$ vs. $S.$ & AR.\% & -0.004 & [-0.025, 0.016] \\
\bottomrule
\end{tabular}
\end{table}

\begin{table}[h]
\centering
\setlength{\tabcolsep}{4pt}
\caption{Paired bootstrap difference confidence intervals for zero-shot editing. Positive values indicate improvement of the main representation.}
\label{tab:paired_edit_stats}
\begin{tabular}{llcc}
\toprule
Comparison & Metric & Difference & 95\% CI \\
\midrule
$S.$ vs. $-$ & Overall ES.\% & +47.10 & [43.80, 50.50] \\
$S.$ vs. $-$ & FE@ES.\% & +46.38 & [41.06, 51.45] \\
$S.$ vs. $-$ & CE@ES.\% & +51.77 & [45.98, 57.56] \\
$S.$ vs. $-$ & DE@ES.\% & +42.91 & [36.36, 49.45] \\
$S.$ vs. $-$ & Overall IR.\% & +29.30 & [26.00, 32.60] \\
\midrule
$S.{+}P.$ vs. $S.$ & Overall ES.\% & +3.60 & [1.20, 6.00] \\
$S.{+}P.$ vs. $S.$ & FE@ES.\% & -2.66 & [-6.04, 0.72] \\
$S.{+}P.$ vs. $S.$ & CE@ES.\% & +4.82 & [0.96, 8.68] \\
$S.{+}P.$ vs. $S.$ & DE@ES.\% & +11.64 & [6.18, 17.09] \\
$S.{+}P.$ vs. $S.$ & Overall IR.\% & +2.30 & [0.30, 4.30] \\
\bottomrule
\end{tabular}
\end{table}

As shown in Table~\ref{tab:paired_edit_stats}, semantic lifting improves zero-shot editing over primitive supervision across all edit categories. Adding parametric lifting provides smaller but targeted gains: the difference is not supported for feature edits, but is supported for consistency and derived-relation edits. This matches the role of parametric lifting as relation-aware supervision rather than generic feature-edit supervision.

\subsection{Breakdown of Zero-shot Intent-Preserving Editing}
Table~\ref{tab:edit_full_breakdown_appendix} provides category-level breakdowns. The main trend is consistent with Table~\ref{tab:edit_overall}: semantic lifting improves overall editability over primitive code, and parametric lifting provides additional gains on relation-sensitive edits that require preserving shared or derived parameters.

\begin{table}[h]
\centering
\small
\setlength{\tabcolsep}{4pt}
\caption{
Detailed results on the edit benchmark. $\Delta S.$ denotes the improvement for semantic lifting. $\Delta P.$ denotes the additional improvement for parametric lifting.
}
\label{tab:edit_full_breakdown_appendix}
\begin{tabular}{@{}lccc ccc cc cc@{}}
\toprule
\multirow{2}{*}{Type}
& \multicolumn{3}{c}{IR.\%$\downarrow$}
& \multicolumn{3}{c}{ES.\%$\uparrow$}
& \multicolumn{2}{c}{$\Delta S.$}
& \multicolumn{2}{c}{$\Delta P.$} \\
\cmidrule(lr){2-4}
\cmidrule(lr){5-7}
\cmidrule(lr){8-9}
\cmidrule(l){10-11}
& $-$ & $S.$ & $S.+P.$
& $-$ & $S.$ & $S.+P.$
& IR & ES
& IR & ES \\
\midrule
Feature
& 28.74 & \textbf{4.11} & \textbf{4.11}
& 48.79 & \textbf{95.17} & 92.51
& +24.63 & +46.38
& +0.00 & -2.66 \\

\quad Param
& 32.50 & \textbf{1.50} & 2.50
& 54.50 & \textbf{98.00} & 97.50
& +31.00 & +43.50
& -1.00 & -0.50 \\

\quad Replace
& 25.23 & 6.54 & \textbf{5.61}
& 43.46 & \textbf{92.52} & 87.85
& +18.69 & +49.06
& +0.93 & -4.67 \\

\midrule
Consistency
& 41.16 & 6.43 & \textbf{3.22}
& 39.23 & 91.00 & \textbf{95.82}
& +34.73 & +51.77
& +3.21 & +4.82 \\

Derived-Relation
& 42.18 & 12.00 & \textbf{7.27}
& 34.18 & 77.09 & \textbf{88.73}
& +30.18 & +42.91
& +4.73 & +11.64 \\
\bottomrule
\end{tabular}
\end{table}

\subsection{Semantic Versus Complexity}
\label{app:semantic_evidence}

Semantic lifting jointly changes feature abstraction, API names, program structure, and target length. To examine whether the benefits of semantic lifting arise solely from shorter targets, we evaluate two additional variants under the same controlled setup. Primitive with semantic comments (\textbf{Prim.+Cmt.}) retains the primitive programs while adding feature-level comments inspired by CADTalk~\cite{yuan2024cadtalk}, without changing executable operations. Semantic with anonymous APIs (\textbf{$S.$-Anon.}) preserves the lifted feature structure and target token length but replaces meaningful API names with fixed anonymous aliases, which are restored before execution. This intervention is related to the semantics-preserving identifier perturbations studied in ReCode~\cite{wang2023recode}.

Both variants are trained independently using the same backbone, data split, and fine-tuning settings, and evaluated on identical test prompts. In \textbf{Prim.+Cmt.}, feature comments are inserted before their corresponding primitive blocks, leaving the original operation sequence unchanged. In \textbf{$S.$-Anon.}, a fixed dataset-wide codebook is used to replace semantic API names, with aliases chosen to preserve target token length under the model tokenizer. The inverse mapping is applied before execution, ensuring that validity and geometry are evaluated using the standard CadQuery APIs. These variants do not fully disentangle semantic content from API familiarity, but provide complementary evidence on whether the observed gains can be explained by shorter targets alone.

\begin{table}[htbp]
% \color{blue}
\centering
% \small
\setlength{\tabcolsep}{4pt}
\caption{Complementary representation variants.
Brackets indicate 95\% bootstrap confidence intervals.
CD is $\times 10^3$, and GT tokens denote mean target-code length.}
\label{tab:semantic_evidence}
\begin{tabular}{lcccc}
\toprule
Rep. & IR.\% $\downarrow$ & Mean CD $\downarrow$ &
Median CD $\downarrow$ & GT tokens \\
\midrule
$-$ & 2.39 [2.09, 2.69] & 1.44 [1.12, 2.18] & 0.14 & 496.76 \\
Prim.+Cmt. & 1.27 [0.87, 1.94] & 0.87 [0.52, 1.22] & \textbf{0.12} & 502.55 \\
$S.$-Anon. & 14.36 [13.69, 15.06] & \textbf{0.38} [0.25, 0.68] & \textbf{0.12} & 339.23 \\
$S.$ & \textbf{0.22} [0.12, 0.33] & 0.46 [0.28, 0.77] & \textbf{0.12} & 339.23 \\
\bottomrule
\end{tabular}
\end{table}

\paragraph{Findings.}
\textbf{Prim.+Cmt.} improves over the primitive representation despite retaining the full primitive generation trajectory with a slightly longer length, showing that explicit semantic guidance is beneficial without target shortening. The further improvement of S. shows that its compact feature-level abstraction provides additional gains. However, although \textbf{$S.$-Anon.} preserves the same lifted program structure and has exactly the same ground-truth code length as S., its IR rises from 0.22\% to 14.36\% and is even worse than \textbf{Prim.+Cmt.}. The substantial degradation indicates that removing meaningful API names may make the representation considerably harder for the model to learn and generate reliably.

\section{Additional Qualitative Comparisons}
\label{app:additional_qualitative}

\subsection{Intent-oriented Text-to-CAD Generation}

\begin{figure}[h]
    \centering
    \includegraphics[width=1\textwidth]{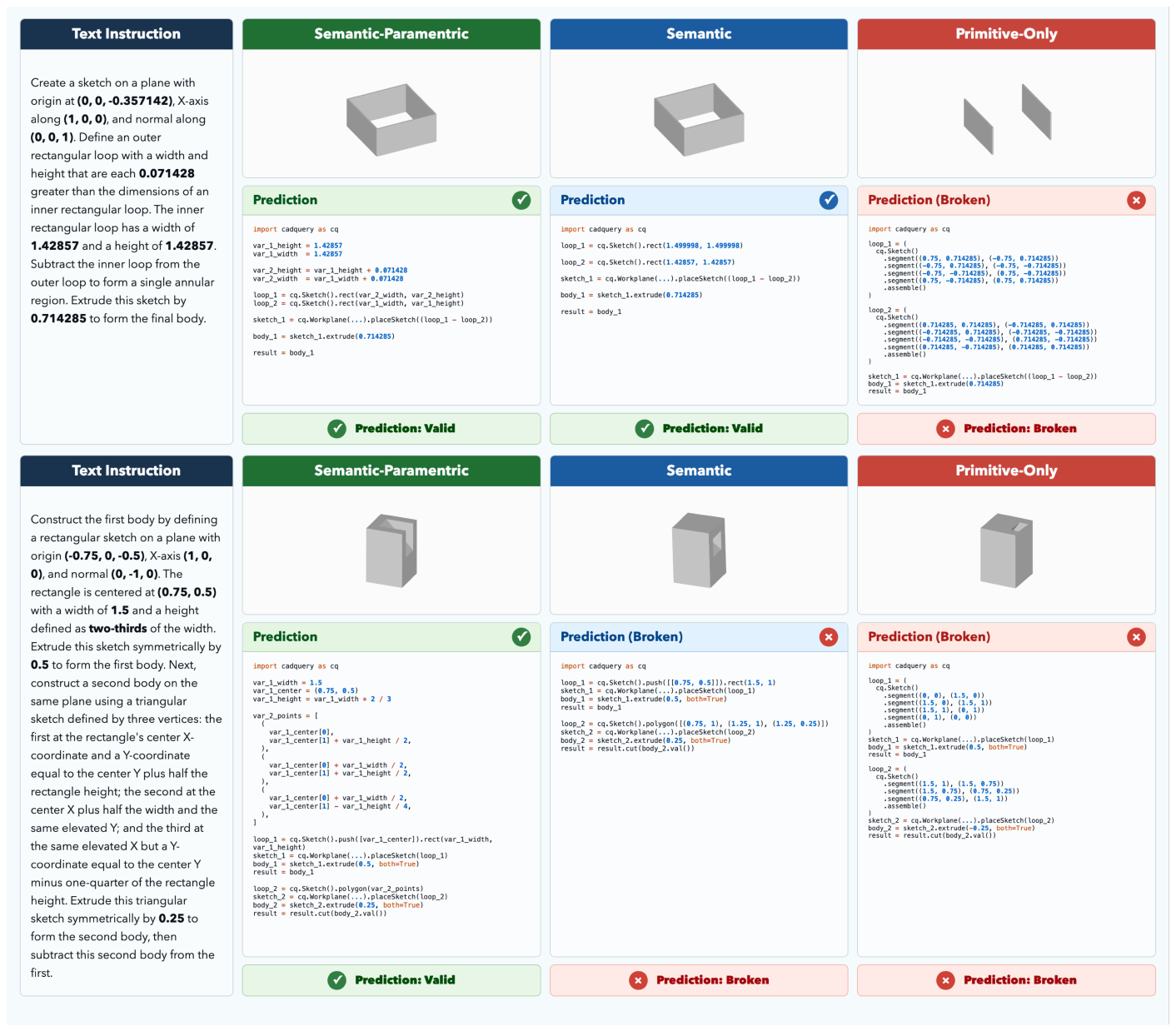}
    \caption{
    Additional Qualitative Comparisons of Intent-oriented Text-to-CAD Generation for primitive-only, semantic, and semantic--parametric supervisions. 
    }
    \label{fig:gen_examples}
\end{figure}

Figure~\ref{fig:gen_examples} presents additional qualitative comparisons of CAD models and corresponding programs generated for primitive-only, semantic, and semantic--parametric supervisions. 

\clearpage
\subsection{Zero-shot Intent-Preserving Editing}

\begin{figure}[h]
    \centering
    \includegraphics[width=1\textwidth]{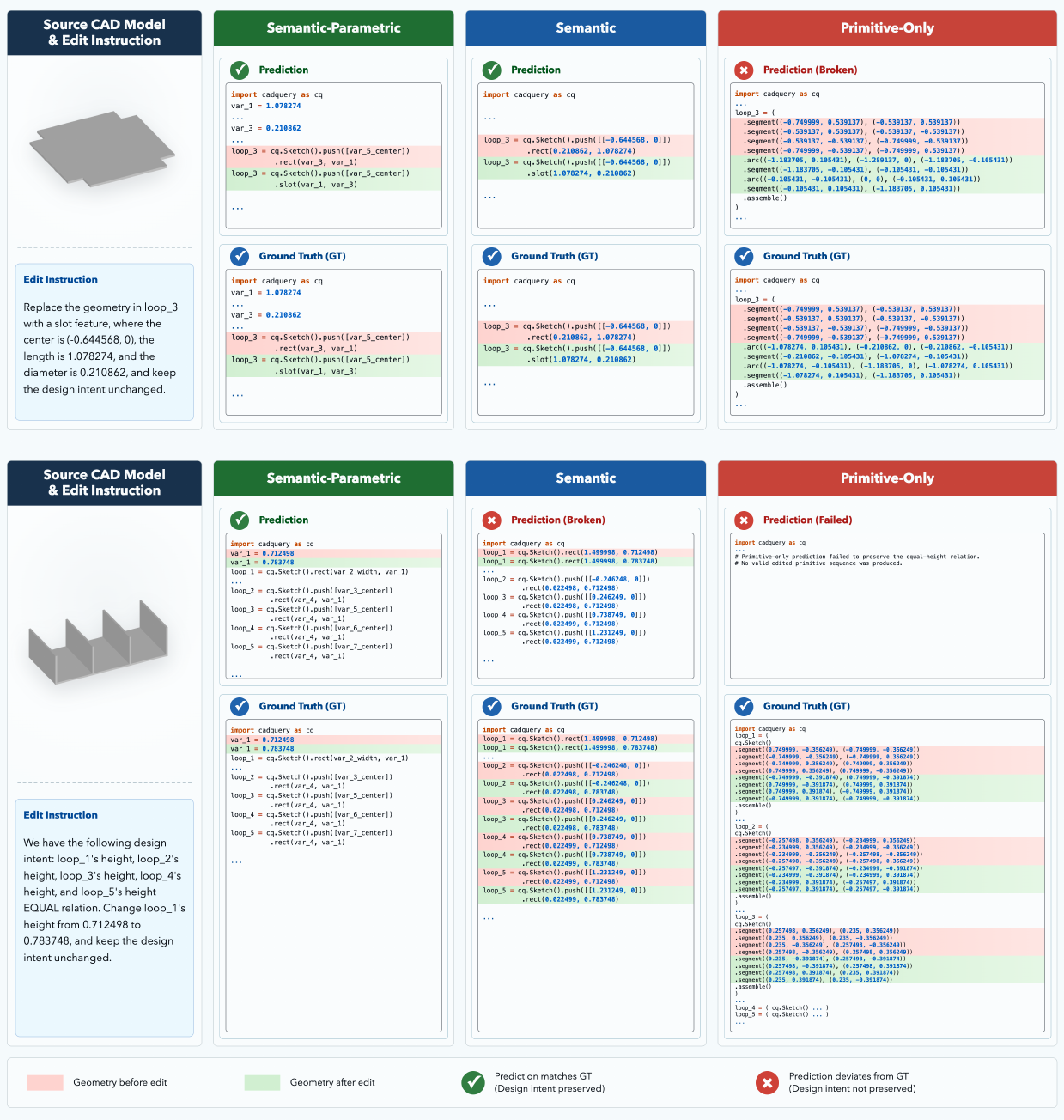}
    \caption{
    Additional Qualitative Comparisons of Zero-shot Intent-Preserving Editing for primitive-only, semantic, and semantic--parametric supervisions. 
    }
    \label{fig:edit_examples}
\end{figure}

Figure~\ref{fig:edit_examples} presents additional qualitative comparisons of edited programs for primitive-only, semantic, and semantic--parametric supervisions. 

\clearpage
\subsection{Additional Qualitative Examples}
\label{app:qualitative_contrasts}

\subsubsection{Text-to-CAD Generation}

\begin{figure}[h]
\centering

\begin{minipage}{0.90\linewidth}
\centering
\setlength{\tabcolsep}{2pt}

\begin{tabular}{@{}cccc@{}}

{\small (a) GT} &
{\small (b) $-$} &
{\small (c) $S.$} &
{\small (d) $S.{+}P.$}
\\[0.3em]

% ============================================================
% Generation Case 1
% ============================================================

\multicolumn{4}{c}{
\parbox{0.98\linewidth}{
\centering\small
\textbf{Prompt:}
\emph{Construct the first body by defining a sketch on a plane
\ldots\ with a circular cutout \ldots\ and unite it with
the existing assembly.}
}}
\\[0.2em]

\includegraphics[width=0.225\linewidth]
{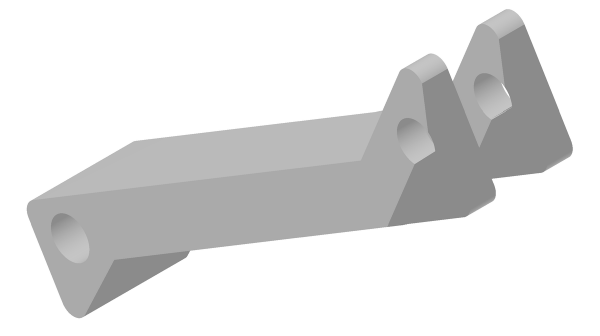} &
\includegraphics[width=0.225\linewidth]
{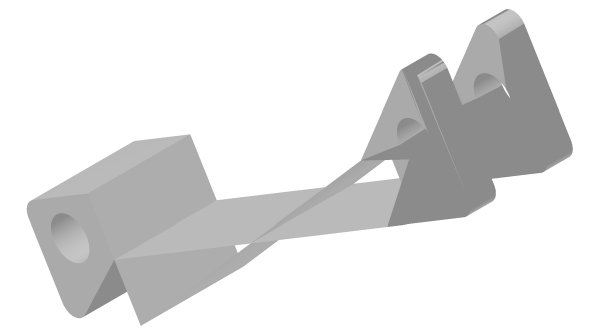} &
\includegraphics[width=0.225\linewidth]
{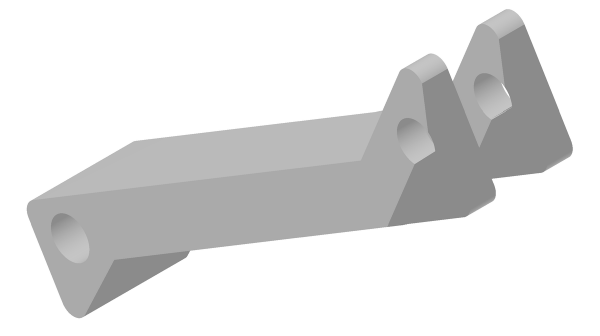} &
\includegraphics[width=0.225\linewidth]
{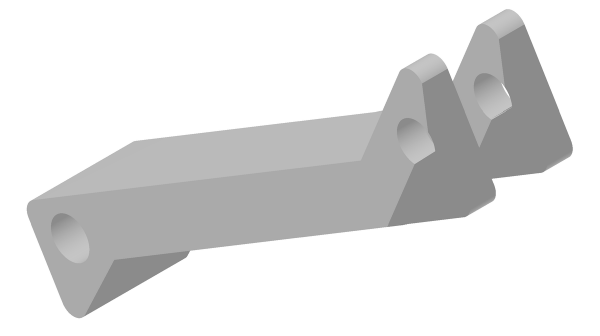}
\\[-0.5em]

{\small Reference} &
{\small \textcolor{red!80!black}{$\times$ Shape Mismatch}} &
{\small \textcolor{green!80!black}{$\checkmark$ Success}} &
{\small \textcolor{green!80!black}{$\checkmark$ Success}}
\\[0.5em]

% ============================================================
% Generation Case 2
% ============================================================

\multicolumn{4}{c}{
\parbox{0.98\linewidth}{
\centering\small
\textbf{Prompt:}
\emph{Construct the first body by sketching a closed loop
on a plane \ldots\ Extrude this sketch symmetrically
\ldots\ subtract it from the remaining geometry.}
}}
\\[0.2em]

\includegraphics[width=0.225\linewidth]
{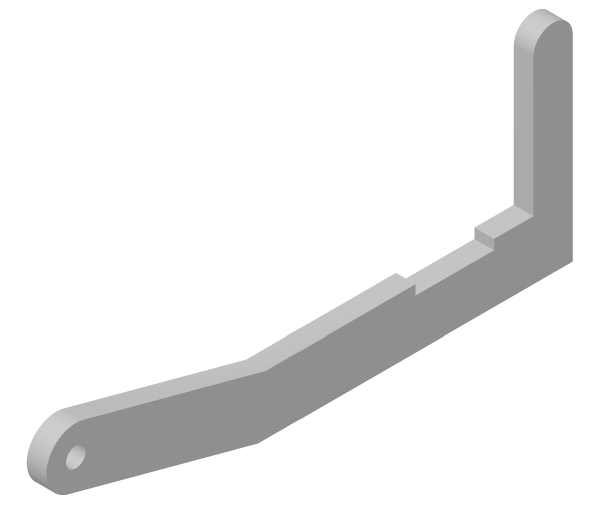} &
\includegraphics[width=0.225\linewidth]
{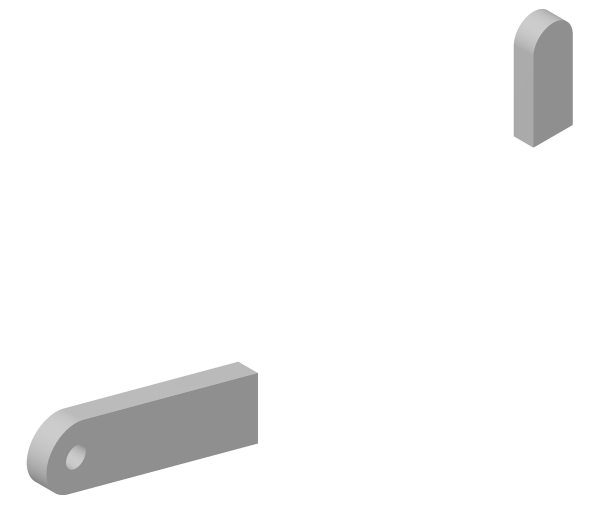} &
\includegraphics[width=0.225\linewidth]
{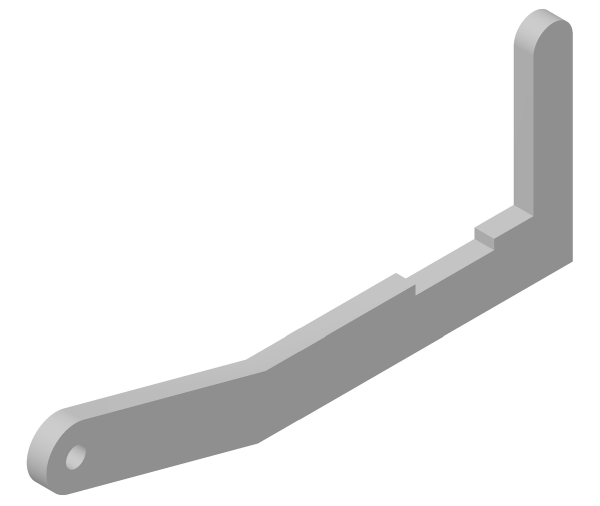} &
\includegraphics[width=0.225\linewidth]
{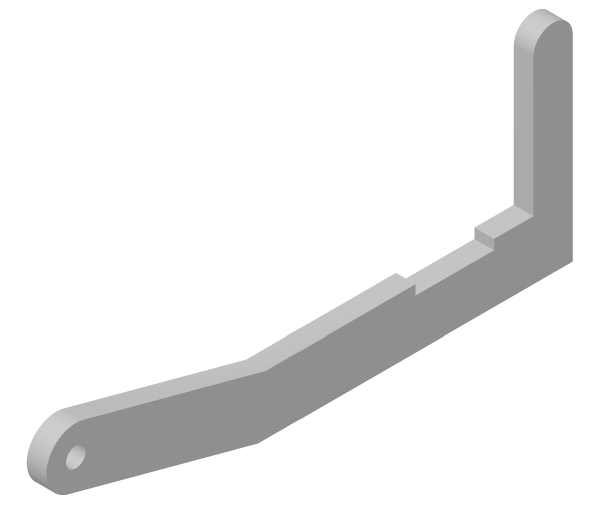}
\\[-0.5em]

{\small Reference} &
{\small \textcolor{red!80!black}{$\times$ Shape Mismatch}} &
{\small \textcolor{green!80!black}{$\checkmark$ Success}} &
{\small \textcolor{green!80!black}{$\checkmark$ Success}}
\\[0.5em]

% ============================================================
% Generation Case 3
% ============================================================

\multicolumn{4}{c}{
\parbox{0.98\linewidth}{
\centering\small
\textbf{Prompt:}
\emph{Construct the first body by sketching a circle with
radius 0.107142 \ldots\ Create the second body by sketching
a regular hexagon \ldots\ Finally, create the sixth body
\ldots}
}}
\\[0.2em]

\includegraphics[width=0.225\linewidth]
{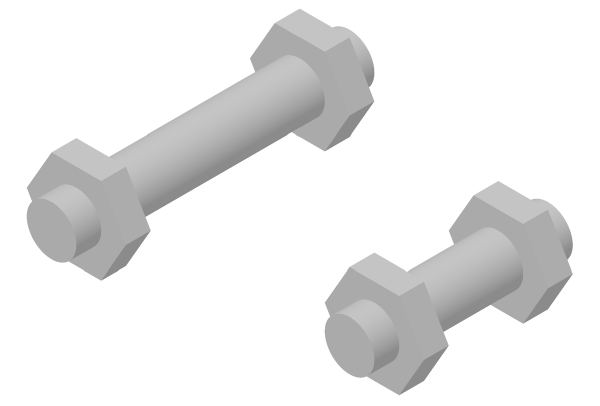} &
\includegraphics[width=0.225\linewidth]
{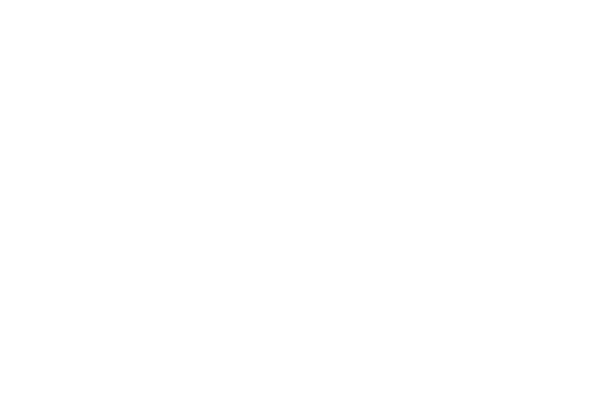} &
\includegraphics[width=0.225\linewidth]
{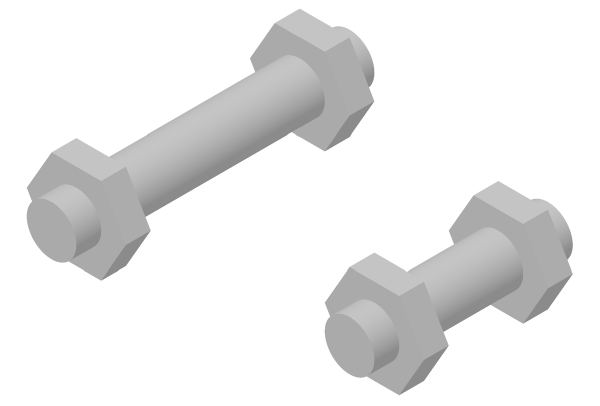} &
\includegraphics[width=0.225\linewidth]
{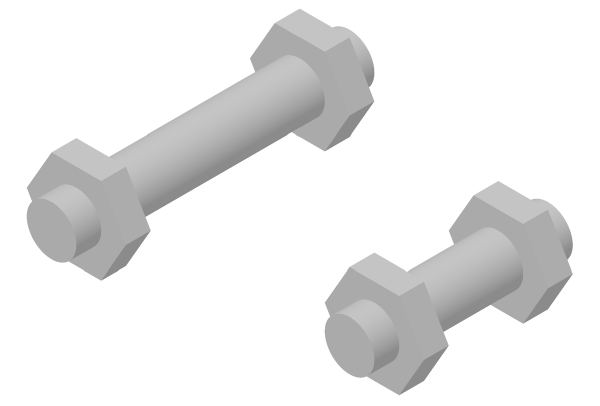}
\\[-0.5em]

{\small Reference} &
{\small \textcolor{red!80!black}{$\times$ Execution Failed}} &
{\small \textcolor{green!80!black}{$\checkmark$ Success}} &
{\small \textcolor{green!80!black}{$\checkmark$ Success}}
\\[0.5em]

% ============================================================
% Generation Case 4
% ============================================================

\multicolumn{4}{c}{
\parbox{0.98\linewidth}{
\centering\small
\textbf{Prompt:}
\emph{Construct the first body by defining a sketch on a plane
\ldots\ This sketch contains a rectangular loop
\ldots\ combined with a slot-shaped region
\ldots\ subtract the resulting body from the first body.}
}}
\\[0.2em]

\includegraphics[width=0.225\linewidth]
{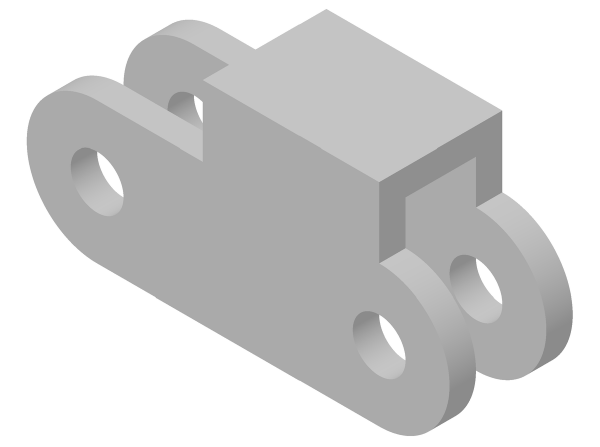} &
\includegraphics[width=0.225\linewidth]
{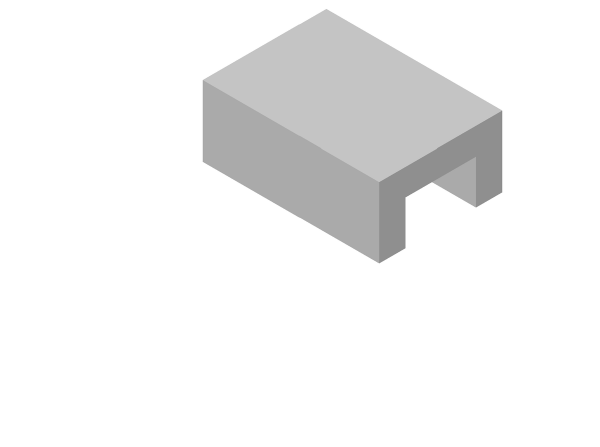} &
\includegraphics[width=0.225\linewidth]
{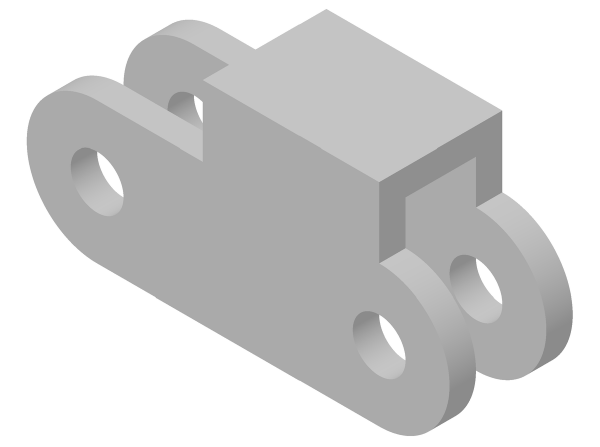} &
\includegraphics[width=0.225\linewidth]
{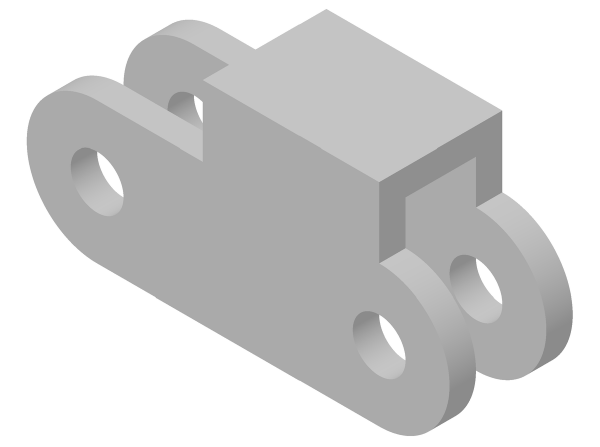}
\\[-0.5em]

{\small Reference} &
{\small \textcolor{red!80!black}{$\times$ Shape Mismatch}} &
{\small \textcolor{green!80!black}{$\checkmark$ Success}} &
{\small \textcolor{green!80!black}{$\checkmark$ Success}}
\\[0.5em]

% ============================================================
% Generation Case 5
% ============================================================

\multicolumn{4}{c}{
\parbox{0.98\linewidth}{
\centering\small
\textbf{Prompt:}
\emph{Construct the first body by defining a sketch on a plane
\ldots\ From this profile, subtract three circular cutouts
\ldots\ Finally, unite the second body with the first body.}
}}
\\[0.2em]

\includegraphics[width=0.225\linewidth]
{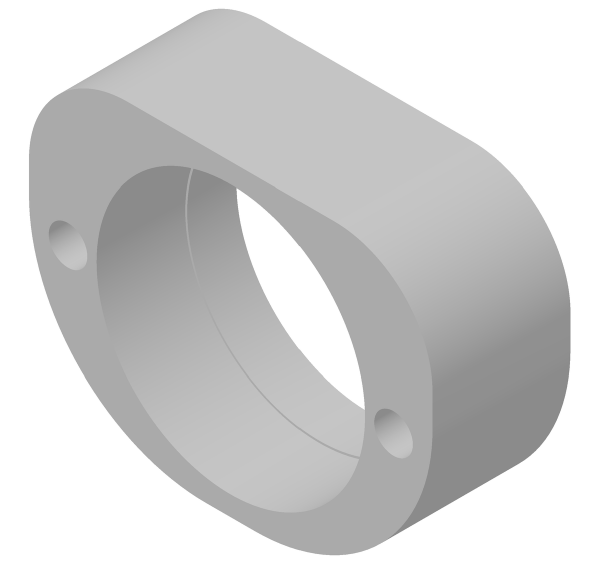} &
\includegraphics[width=0.225\linewidth]
{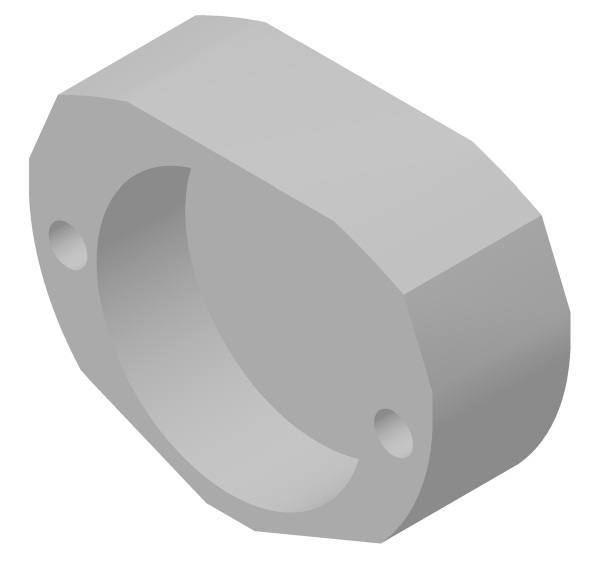} &
\includegraphics[width=0.225\linewidth]
{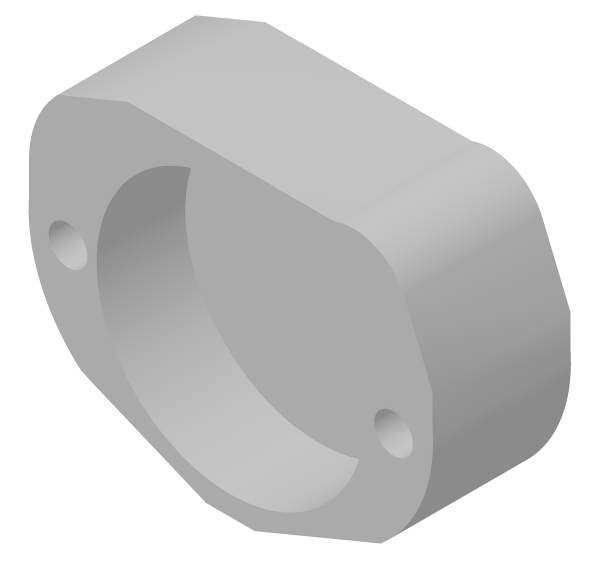} &
\includegraphics[width=0.225\linewidth]
{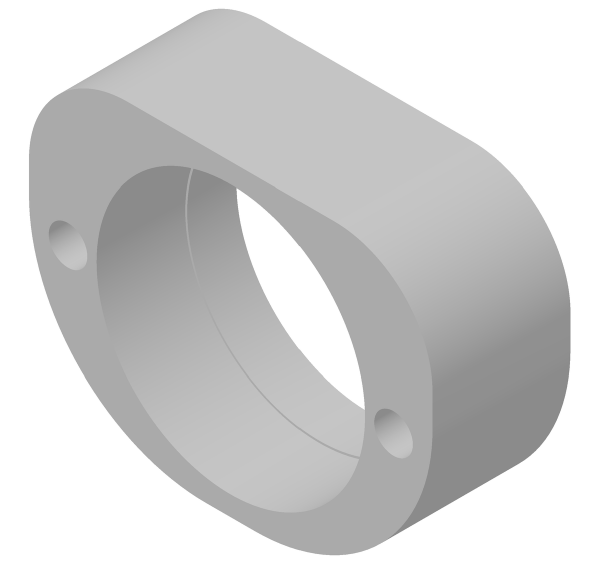}
\\[-0.5em]

{\small Reference} &
{\small \textcolor{red!80!black}{$\times$ Shape Mismatch}} &
{\small \textcolor{red!80!black}{$\times$ Shape Mismatch}} &
{\small \textcolor{green!80!black}{$\checkmark$ Success}}

\end{tabular}
\end{minipage}

\caption{Additional qualitative examples of text-to-CAD generation
across the three representations.}
\label{fig:qual_generation}
\end{figure}

% ============================================================
% Intent-Preserving Editing
% ============================================================
\clearpage
\subsubsection{Intent-Preserving Editing}

\begin{figure}[h]
\centering

\begin{minipage}{0.90\linewidth}
\centering
\setlength{\tabcolsep}{2pt}

\begin{tabular}{@{}cccc@{}}

{\small (a) Source} &
{\small (b) $-$} &
{\small (c) $S.$} &
{\small (d) $S.{+}P.$}
\\[0.3em]

% ============================================================
% Editing Case 1: Mirror Relations
% ============================================================

\multicolumn{4}{c}{
\parbox{0.98\linewidth}{
\centering\small
\textbf{Edit instruction:}
\emph{Change \texttt{loop\_3.center} from
$(-0.227678,-0.227678)$ to $(-0.177678,-0.227678)$,
while preserving its mirror relations with loops 2, 4, and 5.}
}}
\\[0.2em]

\includegraphics[width=0.225\linewidth]
{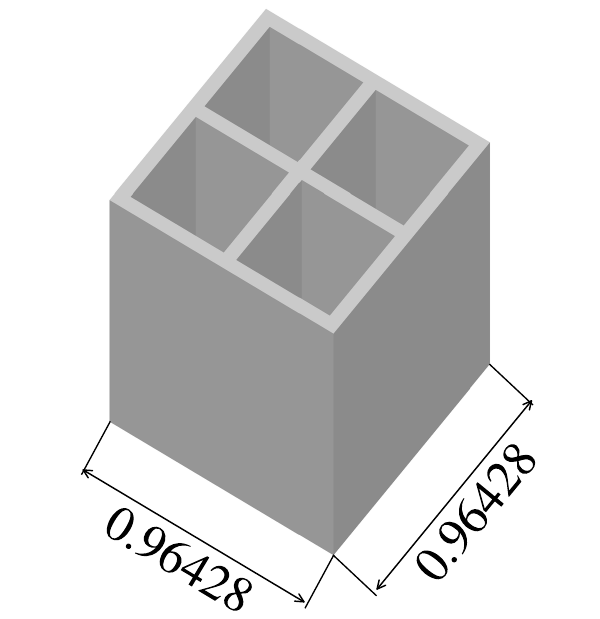} &
\includegraphics[width=0.225\linewidth]
{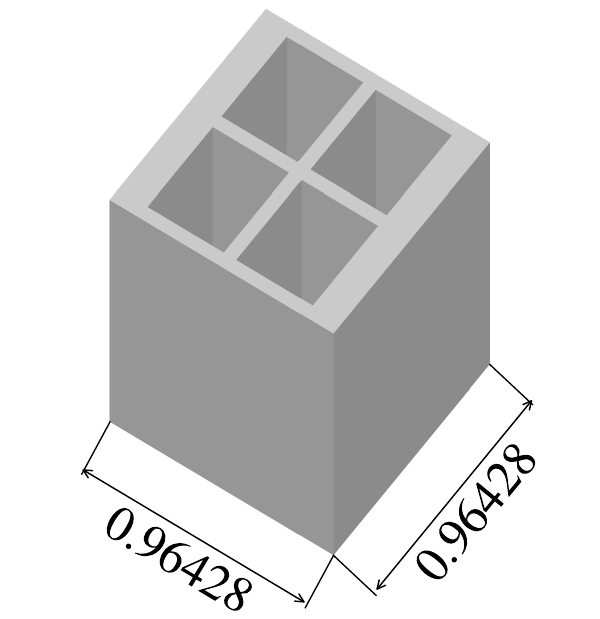} &
\includegraphics[width=0.225\linewidth]
{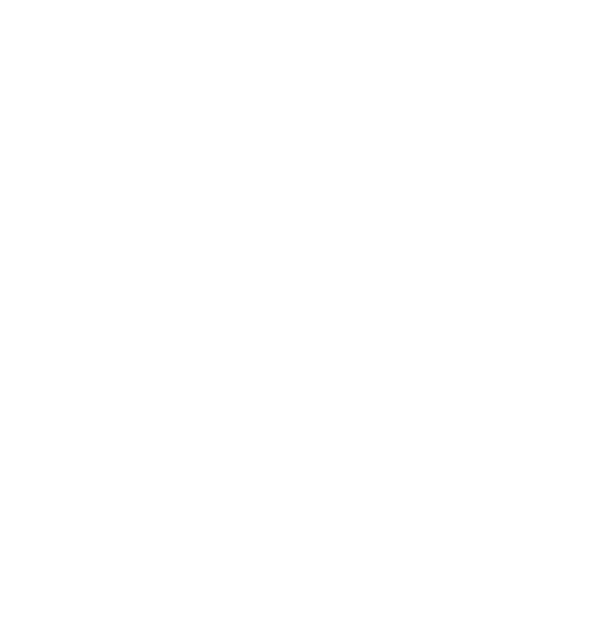} &
\includegraphics[width=0.225\linewidth]
{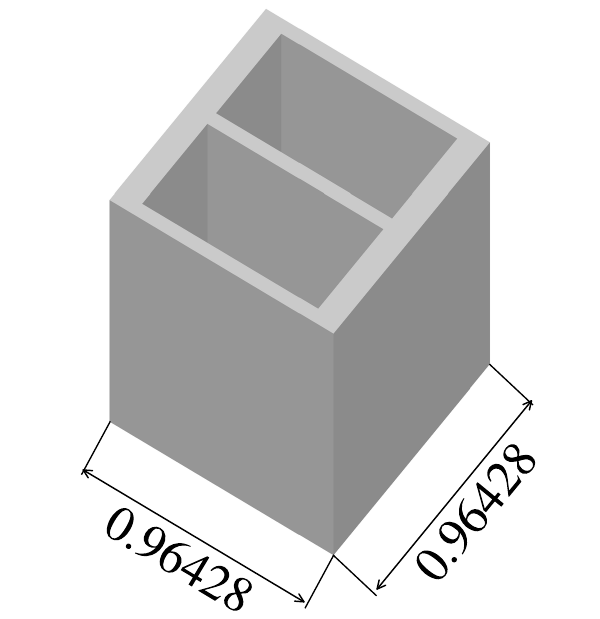}
\\[-0.5em]

{\small $\leftrightarrow$ Mirror Relations} &
{\small \textcolor{red!80!black}{$\times$ Shape Mismatch}} &
{\small \textcolor{red!80!black}{$\times$ Execution Failed}} &
{\small \textcolor{green!80!black}{$\checkmark$ Success}}
\\[0.8em]

% ============================================================
% Editing Case 2: Feature Replacement
% ============================================================

\multicolumn{4}{c}{
\parbox{0.98\linewidth}{
\centering\small
\textbf{Edit instruction:}
\emph{Replace the geometry in \texttt{loop\_1} with a
circle of radius 0.378571, while preserving the design intent.}
}}
\\[0.2em]

\includegraphics[width=0.225\linewidth]
{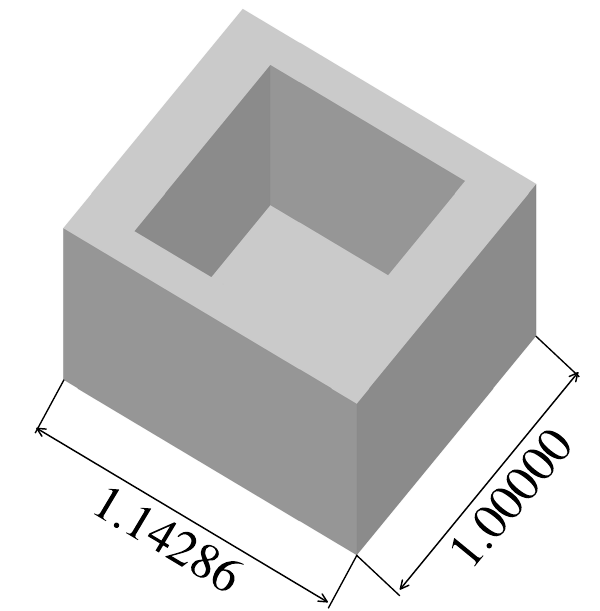} &
\includegraphics[width=0.225\linewidth]
{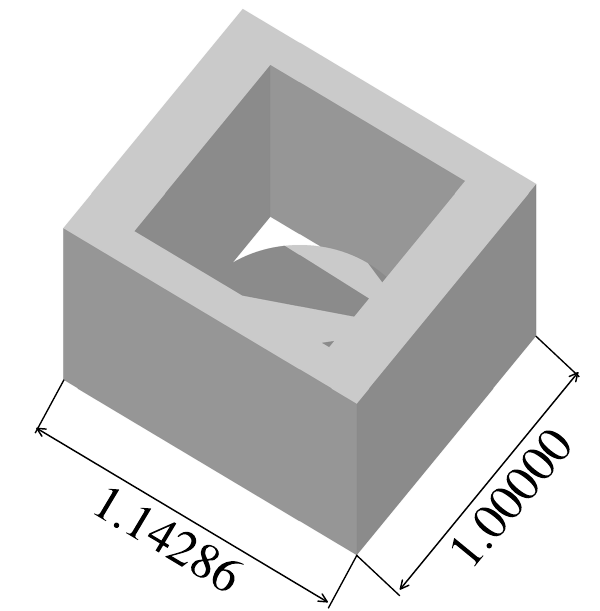} &
\includegraphics[width=0.225\linewidth]
{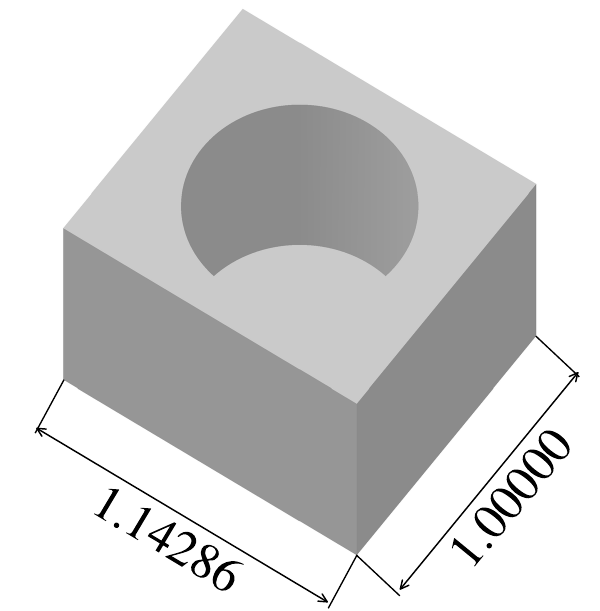} &
\includegraphics[width=0.225\linewidth]
{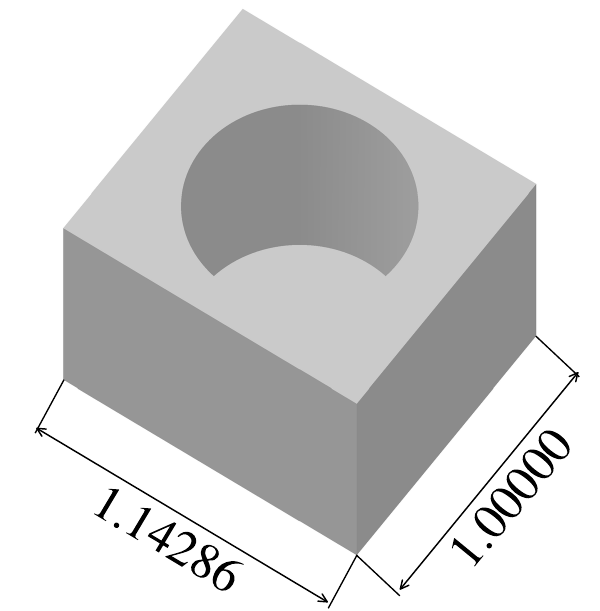}
\\[-0.5em]

{\small Original} &
{\small \textcolor{red!80!black}{$\times$ Shape Mismatch}} &
{\small \textcolor{green!80!black}{$\checkmark$ Success}} &
{\small \textcolor{green!80!black}{$\checkmark$ Success}}
\\[0.8em]

% ============================================================
% Editing Case 3: Fillet Insertion
% ============================================================

\multicolumn{4}{c}{
\parbox{0.98\linewidth}{
\centering\small
\textbf{Edit instruction:}
\emph{Add a fillet feature of radius 0.374999 to the
rectangle in \texttt{loop\_1}, while preserving the design intent.}
}}
\\[0.2em]

\includegraphics[width=0.225\linewidth]
{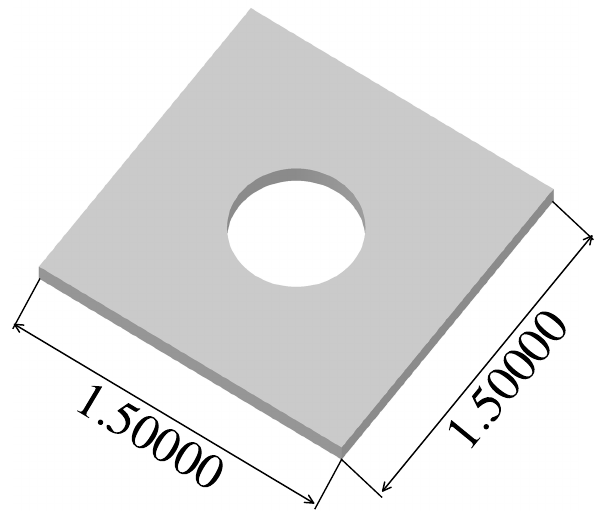} &
\includegraphics[width=0.225\linewidth]
{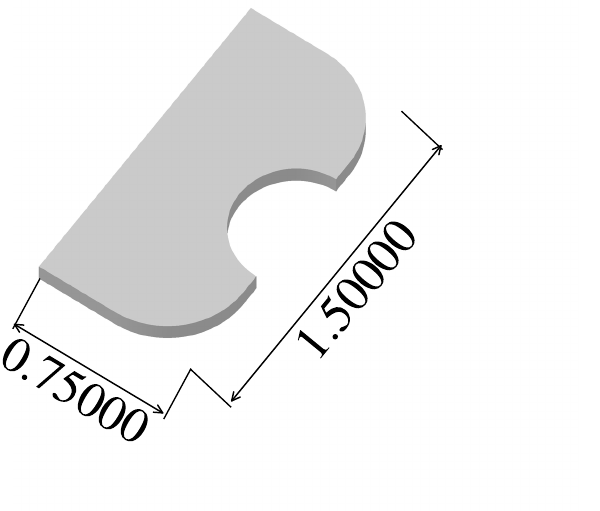} &
\includegraphics[width=0.225\linewidth]
{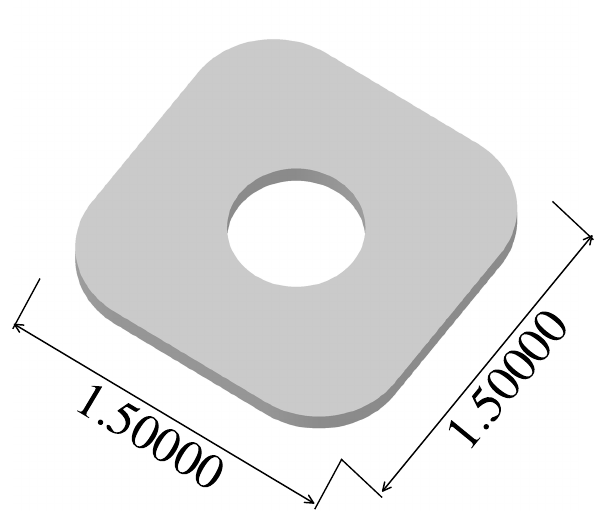} &
\includegraphics[width=0.225\linewidth]
{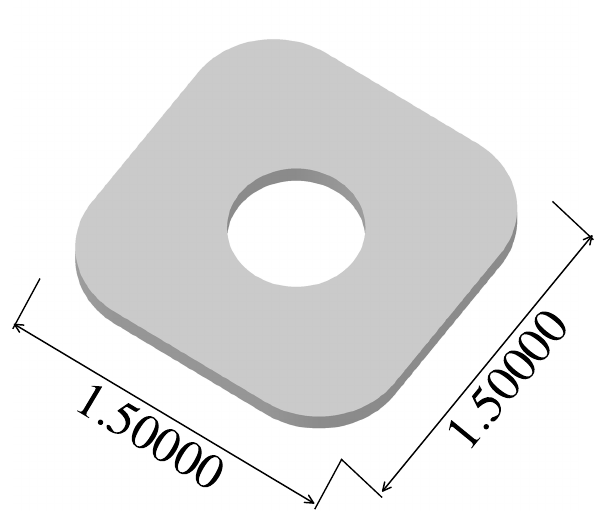}
\\[-0.5em]

{\small Original} &
{\small \textcolor{red!80!black}{$\times$ Shape Mismatch}} &
{\small \textcolor{green!80!black}{$\checkmark$ Success}} &
{\small \textcolor{green!80!black}{$\checkmark$ Success}}
\\[0.8em]

% ============================================================
% Editing Case 4: Equal-Width Relation
% ============================================================

\multicolumn{4}{c}{
\parbox{0.98\linewidth}{
\centering\small
\textbf{Edit instruction:}
\emph{Change \texttt{loop\_2.width} from 0.482991 to
0.53129, while preserving its equal-width relation with
\texttt{loop\_3}.}
}}
\\[0.2em]

\includegraphics[width=0.225\linewidth]
{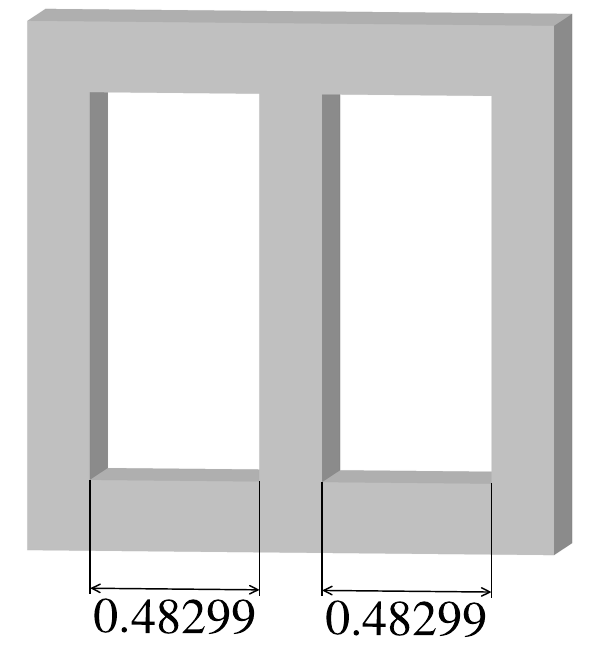} &
\includegraphics[width=0.225\linewidth]
{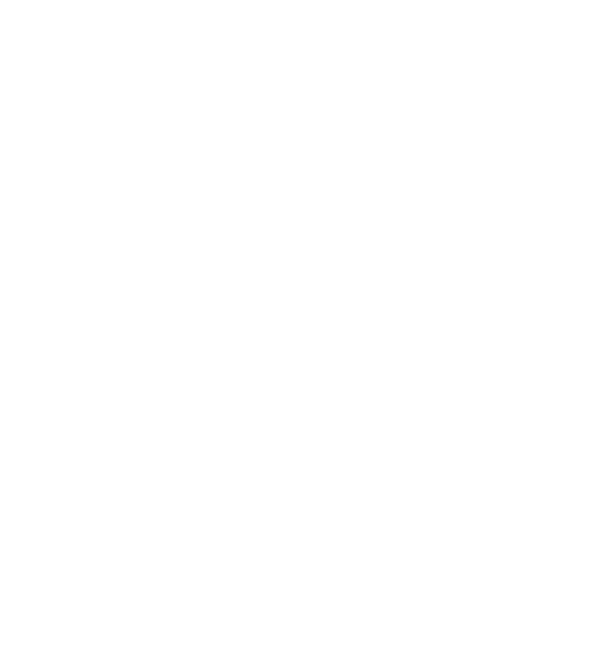} &
\includegraphics[width=0.225\linewidth]
{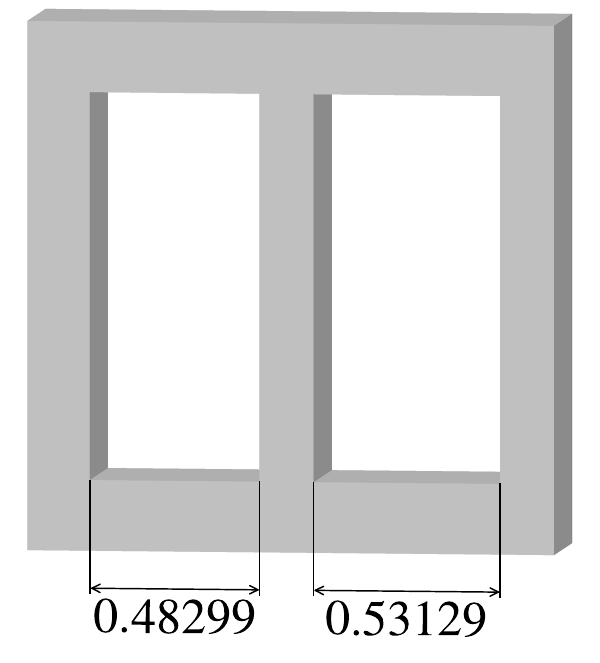} &
\includegraphics[width=0.225\linewidth]
{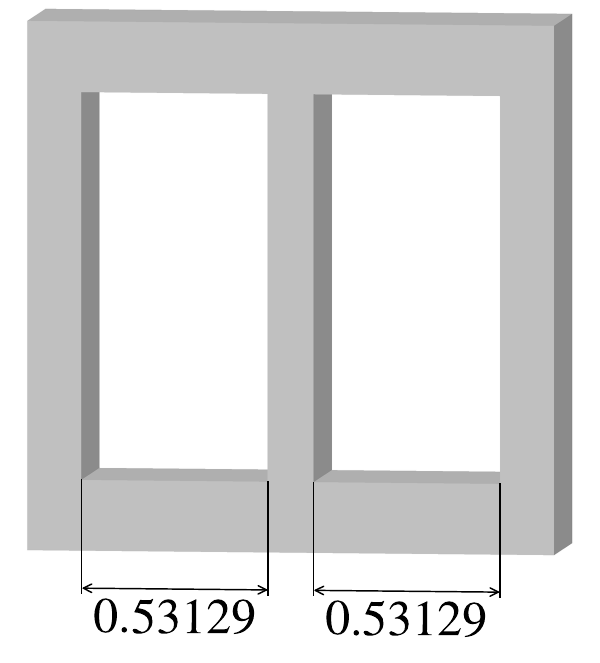}
\\[-0.5em]

{\small $\leftrightarrow$ Equal Widths} &
{\small \textcolor{red!80!black}{$\times$ Execution Failed}} &
{\small \textcolor{yellow!70!black}{\textbf{!} Relation Mismatch}} &
{\small \textcolor{green!80!black}{$\checkmark$ Success}}
\\[0.8em]

% ============================================================
% Editing Case 5: Mirror Relation
% ============================================================

\multicolumn{4}{c}{
\parbox{0.98\linewidth}{
\centering\small
\textbf{Edit instruction:}
\emph{Change \texttt{loop\_5.center} from $(0.529502,0)$
to $(0.579502,0)$, while preserving its mirror relation
with \texttt{loop\_6}.}
}}
\\[0.2em]

\includegraphics[width=0.225\linewidth]
{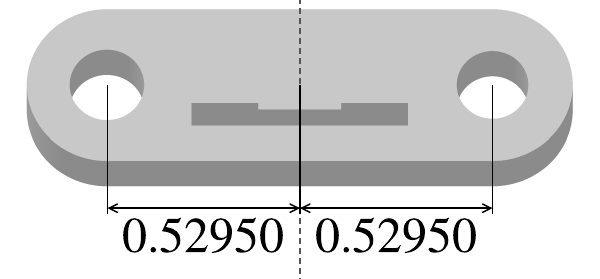} &
\includegraphics[width=0.225\linewidth]
{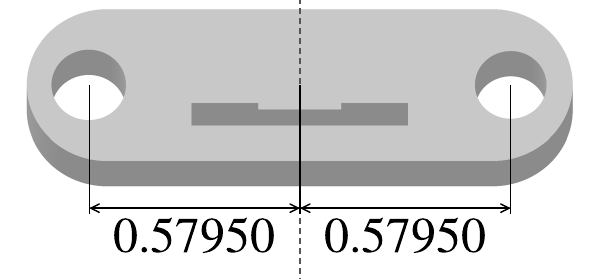} &
\includegraphics[width=0.225\linewidth]
{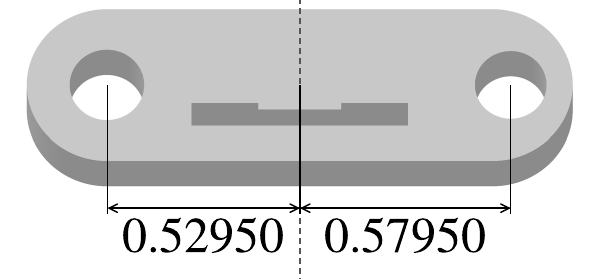} &
\includegraphics[width=0.225\linewidth]
{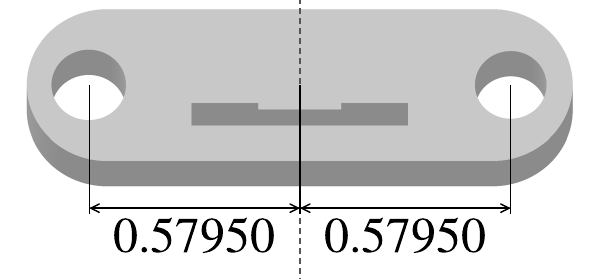}
\\[-0.5em]

{\small $\leftrightarrow$ Mirror Relation} &
{\small \textcolor{green!80!black}{$\checkmark$ Success}} &
{\small \textcolor{yellow!70!black}{\textbf{!} Relation Mismatch}} &
{\small \textcolor{green!80!black}{$\checkmark$ Success}}

\end{tabular}
\end{minipage}

\caption{Additional qualitative examples of intent-preserving editing
across the three representations.}
\label{fig:qual_editing}
\end{figure}

\end{document}